\documentclass{article}

\usepackage[preprint]{neurips_2026}

\usepackage[utf8]{inputenc} 
\usepackage[T1]{fontenc}    
\usepackage{hyperref}       
\usepackage{url}            
\usepackage{booktabs}       
\usepackage{amsfonts}       
\usepackage{nicefrac}       
\usepackage{microtype}      
\usepackage{xcolor}         
\usepackage{natbib}         
\usepackage{graphicx}       
\usepackage{subcaption}     
\usepackage{bm}             
\usepackage{wrapfig}
\usepackage{booktabs}
\usepackage{multirow}
\usepackage{amsmath}
\usepackage{placeins}
\usepackage{adjustbox}

\title{Rethinking Brain Decoding with CLIP: \\The Role of Adversarial Robustness}

\author{%
  Byeongseo Bok$^{1,2}$ \quad
  Futa Waseda$^{2}$ \quad
  Jun Liu$^{2}$ \quad
  Isao Echizen$^{1,2,3}$ \\[0.5em]   
  $^{1}$ The Graduate University for Advanced Studies (SOKENDAI), Kanagawa, Japan\\
  $^{2}$ National Institute of Informatics, Tokyo, Japan \\
  $^{3}$ The University of Tokyo, Tokyo, Japan\\
  \texttt{\{byeongseo-bok, futa-waseda, csjunliu, iechizen\}@nii.ac.jp} \\
}

\begin{document}

\maketitle
\vspace{-0.8em}
  
\begin{abstract}
Brain decoding aims to uncover neural mechanisms by inferring stimulus-related representations from brain signals. In fMRI studies, this is typically achieved by mapping fMRI responses to the latent representations of computational models. Recently, CLIP has become a popular choice for brain decoding due to its rich vision--language embedding space. However, aligning fMRI signals with CLIP representations remains challenging. As CLIP is not explicitly optimized for neural alignment, its representations may capture statistically predictive cues that are only partially reflected in brain activity, limiting decoding performance. In this paper, we investigate whether adversarially robust representations improve neural decoding with CLIP. Adversarial training suppresses non-robust features and promotes more stable, perceptually structured representations, which may better align with brain activity. We evaluate this by fixing the fMRI decoder and varying only the target representation (standard CLIP vs. robust variants) on fMRI-image retrieval and zero-shot classification tasks across NSD and GOD datasets. Empirical results show that this simple change consistently improves task performance and yields stronger alignment across multiple metrics. Attribution analysis further reveals consistently low agreement between standard CLIP and its robust variants, suggesting that adversarial robustness reorganizes feature importance in the visual representation. These findings suggest that the choice of target representation influences neural decoding performance and that adversarial robustness may serve as a useful criterion for brain decoding.
\end{abstract}

\section{Introduction}

With the recent advances of Artificial Neural Networks (ANNs), brain decoding using non-invasive neural modalities such as Functional Magnetic Resonance Imaging (fMRI) has gained significant attention. The goal of brain decoding is to uncover the neural mechanisms by inferring stimulus-related representations from brain activity; to this end, the typical pipeline requires a mapping from fMRI to the human-interpretable semantic representations, such as latent embeddings of ANNs, which serve as proxies for the underlying stimuli.

For this semantic proxy, early approaches utilized engineered image features like Gabor wavelet patches~\cite{miyawaki2008visual}. With the advent of deep learning, end-to-end feature extractors became widely used~\cite{horikawa2017generic,shen2019deep}. Recently, large-scale vision--language pretrained models such as CLIP~\cite{radford2021learningtransferablevisualmodels} have emerged as effective target representations due to their semantically rich embedding space, and are now broadly adopted in brain decoding~\cite{liu2025surveyfmribasedbraindecoding, wang2025representation, Li_2025neuraldiffuser, wang2024mindbridge, scotti2023reconstructingmindseyefmritoimage, gong2024neuroclips, brainclip, efird2025finding, wu2025bridgingvisionbraingapuncertaintyaware}.  

Despite its popularity, aligning fMRI signals with CLIP representations remains challenging. Recent works~\cite{chen2025bridging,vafaei2026brain} have shown that neural decoding performance with CLIP can be further improved via a biologically constrained pipeline, such as adding a visual attention mechanism~\cite{desimone1995neural} or enforcing CLIP embeddings with geometric similarity to fMRI responses~\cite{vafaei2026brain}, suggesting that the native CLIP representation space may be suboptimal for neural decoding. Moreover, Shirakawa et al.~\cite{shirakawa2025spurious} indicate that visually plausible reconstructions can arise without faithfully capturing underlying neural representations, highlighting that the decoding pipeline can rely on visual features that are predictive yet less aligned with human neural representations. At the representation level, since CLIP is optimized for web-scale image--text contrastive learning rather than neural alignment, its embeddings may capture non-robust~\cite{ilyas2019adversarialexamplesbugsfeatures, cui2024robustness} or shortcut-driven features~\cite{geirhos2020shortcut, wang2024soberlookrobustnessclips} that are predictive for image--text matching but not necessarily well-aligned with brain activity. Together, these findings suggest that improving brain decoding may require revisiting the properties of the target representation itself, beyond architectural or pipeline-level modifications.

Therefore, we hypothesize that adversarially robust CLIP variants can improve neural alignment, defined as the mapping from fMRI responses to CLIP latent representations. Since human neural representations are reported to be relatively invariant to small perturbations and less reliant on non-robust cues~\cite{veerabadran2023subtle}, such features may not be reflected in fMRI responses. Adversarial training has been shown to reduce reliance on non-robust features that are less aligned with human perception~\cite{ilyas2019adversarialexamplesbugsfeatures, tsipras2019robustnessoddsaccuracy}, which have been associated with shortcut-like behavior~\cite{li2023adversarial} and spurious correlations~\cite{zhang2022causaladvadversarialrobustnesslens}. As such, this motivates the central question: ``Do adversarially robust representations improve alignment in neural decoding?''

In this paper, we propose adversarial robustness as a criterion for selecting target representations in brain decoding. Adversarially trained models have been shown to exhibit multiple properties that are more human-aligned than standard models. These include more stable and perceptually meaningful visual features~\cite{ilyas2019adversarialexamplesbugsfeatures,engstrom2019adversarialrobustnesspriorlearned}, more interpretable gradients~\cite{tsipras2019robustnessoddsaccuracy,ross2018improving, kim2019bridgingadversarialrobustnessgradient}, more concise and sparse attribution maps~\cite{chalasani2020conciseexplanationsneuralnetworks, etmann2019connectionadversarialrobustnesssaliency, gong2025boosting}, and increased reliance on structured visual features such as shape-bias~\cite{geirhos2021partialsuccessclosinggap,subramanian2023spatial, gavrikov2023extendedstudyhumanlikebehavior}. This motivates their potential to improve alignment between model representations and neural signals. Empirically, we observe that adversarially trained CLIP variants consistently achieve stronger alignment across multiple metrics and improved fMRI-image retrieval accuracy. Here, we summarize our contribution as follows:

\begin{itemize}
    \item We show that adversarially trained CLIP variants, FARE~\cite{schlarmann2024robust} and TeCoA~\cite{mao2022understanding}, improve alignment between fMRI responses and model representations as measured by Pearson correlation, cosine similarity, and Representational Similarity Analysis (RSA)~\cite{kriegeskorte2008representational}, yielding significantly higher fMRI-image retrieval accuracy on the NSD dataset.
    \item We further show that FARE and TeCoA improve zero-shot performance on fMRI-image retrieval and fMRI-text classification on the GOD dataset, a dataset carefully designed for zero-shot settings with non-overlapping train-test semantics~\cite{horikawa2017generic}.
    \item Finally, using an attribution-based analysis framework introduced in~\cite{palazzo2020decoding}, we show that adversarially trained models exhibit low agreement with standard CLIP in attribution patterns, yielding spatially concentrated and distinct patterns.   
\end{itemize}

\begin{figure}
    \centering
    \includegraphics[width=\linewidth]{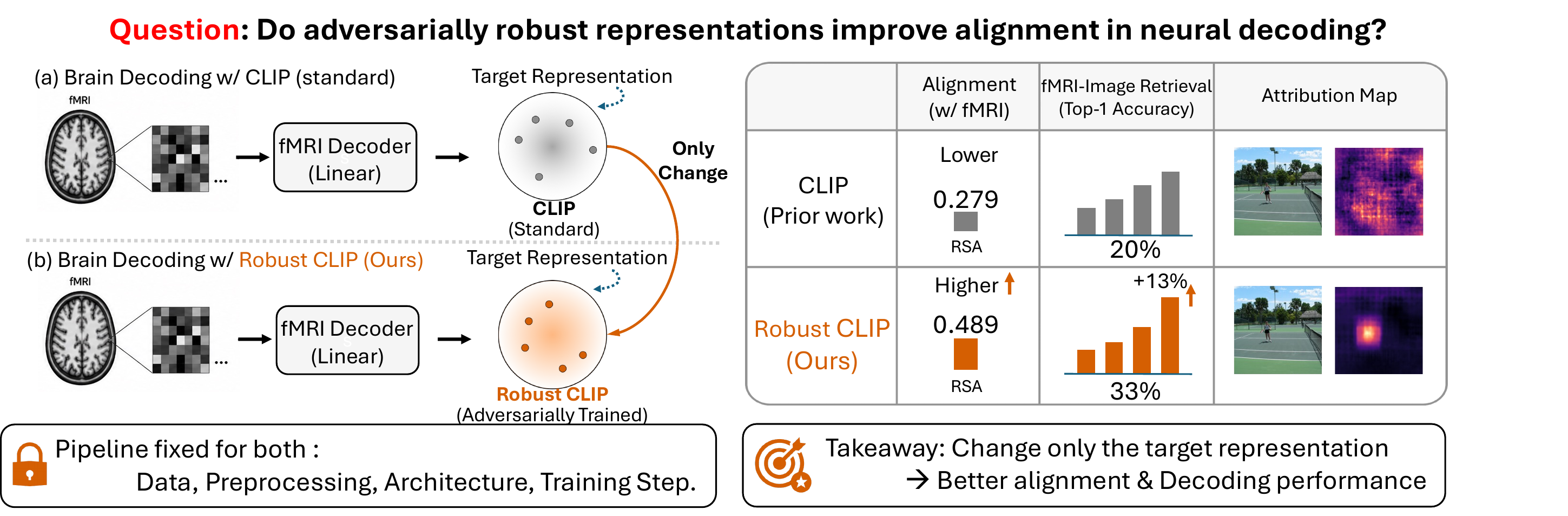}
    \caption{A summary of our work: changing target representation to an adversarially robust one improves neural decoding with CLIP. Empirical results support increased neural alignment and fMRI-image retrieval performance, while attribution maps further reveal qualitatively different image feature usage.}
    \label{fig:teaser_fig}
\end{figure}

\section{Background and Setup}
\subsection{Related Works}


\textbf{Brain Decoding with CLIP.}
Since its remarkable zero-shot generalization capability~\cite{radford2021learningtransferablevisualmodels}, both image~\cite{brainclip, efird2025finding, chen2025bridging, gong2024neuroclips, lin2022mindreaderreconstructingcomplex}and text latent embeddings~\cite{Li_2025neuraldiffuser,scotti2023reconstructingmindseyefmritoimage, takagi2023high} of CLIP have been widely adopted as target representation spaces for brain decoding. Despite its popularity, Shirakawa et al.~\cite{shirakawa2025spurious} show that CLIP-driven reconstructions can be spurious under distribution shifts, raising concerns on reliability. Beyond pipeline-wise concern, several works explored modifying the CLIP representation space itself. Chen et al.~\cite{chen2025bridging} showed that simulating visual attentions to extract selective semantics from CLIP embeddings can improve decoding quality, while Vafaei et al.~\cite{vafaei2026brain} demonstrated that aligning the geometry of pretrained embeddings, including CLIP, with fMRI responses improves decoding performance. These works exhibited current limitations on brain decoding with native CLIP representation and presented engineering solutions to modify the representation space. In contrast, our work does not transform the embedding space, but instead investigates how intrinsic properties of the representation, specifically adversarial robustness, affect brain decoding performance. 

\vspace{3pt}\textbf{Adversarial Robustness.}
Adversarial training~\cite{goodfellow2014explaining,madry2019deeplearningmodelsresistant} has been extensively studied for improving adversarial robustness by suppressing non-robust features~\cite{ilyas2019adversarialexamplesbugsfeatures, tsipras2019robustnessoddsaccuracy} which are associated with shortcut-like behavior and part of spurious correlations in previous works~\cite{li2023adversarial, zhang2022causaladvadversarialrobustnesslens}. Beyond robustness, previous works have shown that adversarially trained models exhibit
more structured representations, including human-interpretable  gradients~\cite{tsipras2019robustnessoddsaccuracy,ross2018improving, kim2019bridgingadversarialrobustnessgradient}, concise and spatially-concentrated attribution maps~\cite{chalasani2020conciseexplanationsneuralnetworks, etmann2019connectionadversarialrobustnesssaliency, gong2025boosting}, and increased reliance on more structured features like shape-bias~\cite{geirhos2021partialsuccessclosinggap,subramanian2023spatial, gavrikov2023extendedstudyhumanlikebehavior}. Several studies have also shown that incorporating biological constraints into models can improve robustness and perceptual alignment~\cite{dapello2020simulating, li2019learningbrainsregularizemachines, guo2024limitedconsistentgainsadversarial}, connecting adversarial robustness with brain-like representations~\cite{nonaka2021brain,mineault2025neuroaiaisafety}. However, it remains unclear whether the converse holds—namely, whether adversarially robust representations are better aligned with neural signals in brain decoding.

\subsection{Experiment Setups }
\subsubsection{Dataset}
\textbf{NSD dataset.}
Natural Scene Dataset (NSD)~\cite{allen2022massive} is currently the largest open-source vision--fMRI dataset. NSD contains 73,000 static images cropped from the MSCOCO dataset~\cite{lin2015microsoftcococommonobjects} with corresponding fMRI activation data attained from 8 different subjects viewing these images during multiple fMRI trials. In our work, we followed the data preprocessing pipeline introduced in ~\cite{efird2025finding}, using denoised beta responses and selecting voxels based on noise ceiling criteria. Following prior work, we evaluate on the standard NSD test split~\cite{allen2022massive, wang2024mindbridge, brainclip} and report results on subjects 1, 2, 5, and 7, which have full trial repetitions. More detailed preprocessing and data handling procedures are provided in the \textbf{Appendix~\ref{sec:ap_dataset}}.

\vspace{3pt}\textbf{GOD dataset.}
Generic Object Decoding Dataset (GOD)~\cite{horikawa2017generic} was introduced with the motivation for testing the generalization ability of models to novel stimuli (zero-shot). GOD contains 1250 images collected from ImageNet~\cite{krizhevsky2012imagenet} and corresponding beta data from 5 different subjects. The last 50 images were used for test data, and 35 repeated trial betas for these test data were averaged to generate a single beta file in our experiment. 

\subsubsection{Problem Setting}
\label{train_scheme}

The typical fMRI-based brain decoding pipelines include: 1) a \textbf{decoder} that maps fMRI responses to the latent representations of a computational model. 2) a \textbf{generator} that reconstructs the stimuli (e.g., images) from these representations. While recent work has often focused on improving reconstruction quality by enhancing the generator component of the pipeline~\cite{takagi2023high, Li_2025neuraldiffuser, scotti2023reconstructingmindseyefmritoimage, liu2025surveyfmribasedbraindecoding}, our approach isolates the decoder, aiming to study the fidelity of the mapping between fMRI responses and target representation without confounding effects from downstream generation. 

\vspace{3pt}\textbf{Training Objective.} 
For the decoder, we train a linear model $d(X;\theta)= \hat{Y}$ following ~\cite{efird2025finding}. The model maps fMRI voxel data $X=[x_i, ..., x_n], \, x_i \in \mathbb{R}^v$ to predicted embeddings $\hat{Y}=[\hat{y}_1, ..., \hat{y}_n], \, \hat{y}_i \in \mathbb{R}^{768}$, where $n$ and $v$ denote the number of samples and subject-specific voxels, respectively. Here, $\hat{Y}$ approximates the target embeddings $Y_M$ from model $M$.

For the main experiments, we optimize the decoder using the InfoNCE loss~\cite{oord2019representationlearningcontrastivepredictive}:
\begin{equation}
        Contrast(A, B) = - \frac{1}{N} \sum_{i=1}^{N}\log\left(\frac{\exp(\bm{a}_i \cdot \bm{b}_i /\tau)}{\sum_{j=1}^{N}\exp(\bm{a}_i \cdot \bm{b}_j/\tau)}\right)
\end{equation}
\begin{equation}
        L_{\mathrm{infoNCE}}(A,B) = \frac{1}{2}[Contrast(A,B)+Contrast(B,A)]
\end{equation}
where $A = \{\bm{a}_i\}_{i=1}^N$ and $B = \{\bm{b}_i\}_{i=1}^N$ denote batches of paired embeddings, $\bm{a}_i$ and $\bm{b}_i$ are the $i$-th embeddings in each batch, and $\tau$ is a temperature hyperparameter. In our setting, the loss is applied to the predicted embeddings $\hat{Y}$ and target embeddings $Y_M$, i.e., $L(\hat{Y},Y_M)=L(d(X;\theta), Y_M)$ where $L$ denotes $L_{\mathrm{infoNCE}}$.

While contrastive training has been shown to improve fMRI-image retrieval accuracy~\cite{brainclip, efird2025finding}, we additionally report results using ridge regression, a standard baseline in brain decoding~\cite{kriegeskorte2019interpretingencodingdecodingmodels}:
\begin{equation}
\hat{B} = \arg\min_{B}
\left\{
\sum_{i=1}^{n} \| y_i - \beta_0 - x_i^\top B \|_2^2
+ \lambda \|B\|_F^2
\right\}
\end{equation}
where $y_i$ denotes the $i$-th target embedding from $Y_M$, while $B \in \mathbb{R}^{v \times 768}$ and $\beta_0 \in \mathbb{R}^{768}$ denote the regression weights and intercept, respectively. The intercept is not regularized.

\vspace{3pt}\textbf{Target Representation Models.}
While many existing works~\cite{chen2025bridging, vafaei2026brain} improve brain decoding by engineering target representations, we instead study how the adversarial robustness of representations affects decoding performance.

To this end, we adopt a controlled setting in which only the target representation is varied while other components remain fixed.  We consider two adversarially robust CLIP variants, TeCoA~\cite{mao2022understanding} and FARE~\cite{schlarmann2024robust}, enabling us to examine how adversarial robustness in the target space influences decoding performance through a simple substitution of target representations. 

TeCoA is a robust CLIP variant in which the visual encoder is adversarially fine-tuned under text-guided supervision. In contrast, FARE employs an unsupervised adversarial fine-tuning scheme that does not rely on text supervision, while maintaining strong zero-shot capabilities.

All models were implemented using publicly available HuggingFace model checkpoints. We use a consistent ViT-L/14 backbone across CLIP, TeCoA, and FARE to ensure a controlled comparison. For each robust variant, we evaluate two levels of adversarial strength: FARE-2 / FARE-4 and TeCoA-2 / TeCoA-4, where the suffix indicates the perturbation budget $\epsilon$ used during adversarial training (e.g., $\epsilon=2/255$ for FARE-2). 

In all experiments, the decoder architecture and training procedure are fixed, and only the target representation model is varied. We refer to decoders trained to predict CLIP, TeCoA, and FARE representations as the CLIP decoder, TeCoA decoders, and FARE decoders, respectively.

\vspace{3pt}\textbf{Evaluation.} We used standard metrics from brain decoding literature. The main metric is top-k accuracy (Recall@k) previously addressed in ~\cite{efird2025finding, brainclip} for the fMRI-image retrieval task, both in the NSD and GOD datasets. In case of the NSD dataset, we present the results from subjects 1, 2, 5, and 7 in the main document for clarity, while the results are consistent across subjects and summarized in the \textbf{Appendix~\ref{sec:ap_alignment_retrieval}}. 

We additionally provide other measures of neural alignment between fMRI signals and target model representations, including Pearson correlation, cosine similarity, and Representational Similarity Analysis (RSA)~\cite{kriegeskorte2008representational} on the NSD test dataset. We evaluate this alignment at two levels. First, as decoder-mediated alignment of fMRI responses, we measure Pearson correlation and cosine similarity between predicted embeddings and target model representations (CLIP, FARE, and TeCoA). Second, we assess representational similarity analysis (RSA) directly between fMRI responses and each target model representation. 

\section{Results}
To examine whether adversarially robust representations improve alignment in neural decoding, in this section, we provide the experimental results based on the training and evaluation scheme described in section~\ref {train_scheme}. We observe a consistent trend that robust models achieve improved decoding performance both on the NSD and GOD datasets.

\subsection{Decoders with Adversarially Trained Models Improve Brain Decoding on the NSD Dataset}

\begin{figure}
  \centering
  \includegraphics[height=0.20\textheight,width=0.85\linewidth]{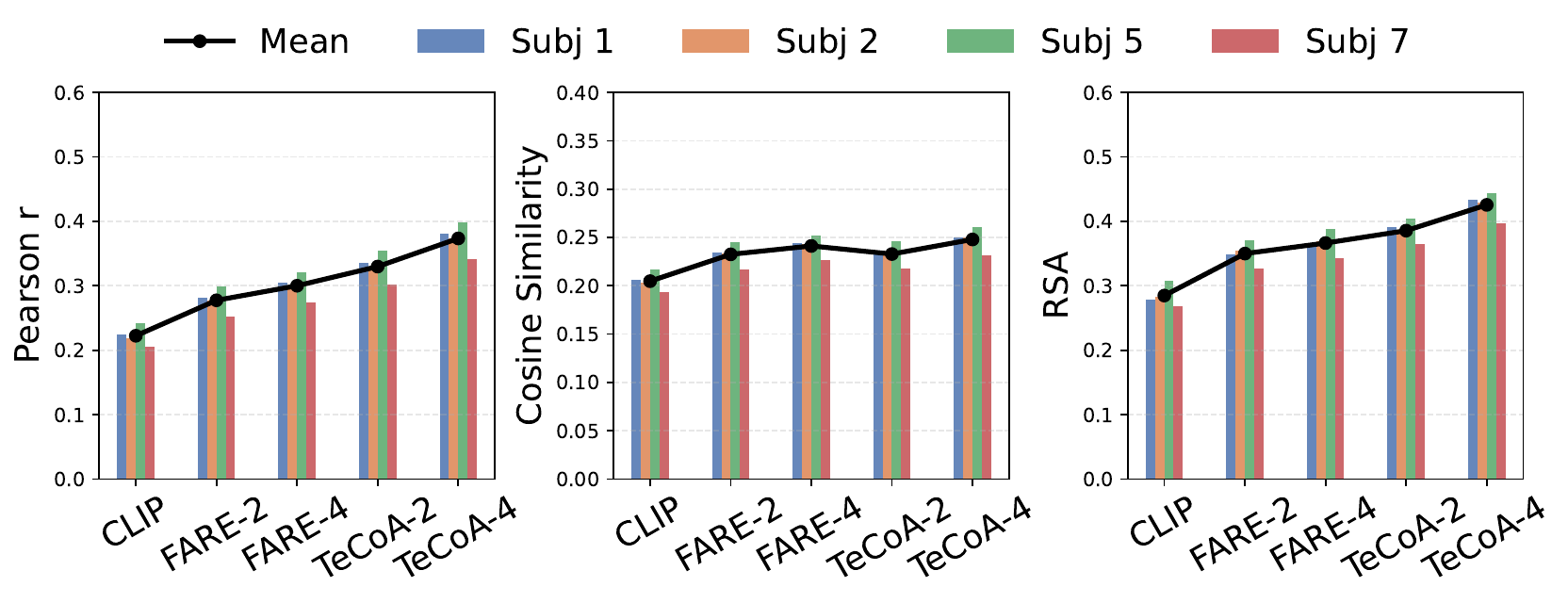}
  \caption{Comparison of alignment metrics between different models on the NSD dataset. All the measures are visualized with subjects 1, 2, 5, 7, and their mean. Adversarially trained variants achieved higher scores in all the metrics.}
    \label{alignment:a}
\end{figure}


\begin{figure}[t]
  \centering

  \includegraphics[width=0.8\linewidth]{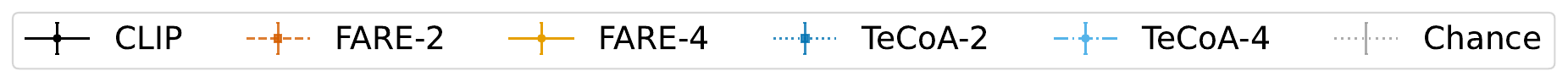}

  \vspace{3pt}

  \begin{subfigure}[t]{0.495\linewidth}
    \centering
    \includegraphics[width=\linewidth]{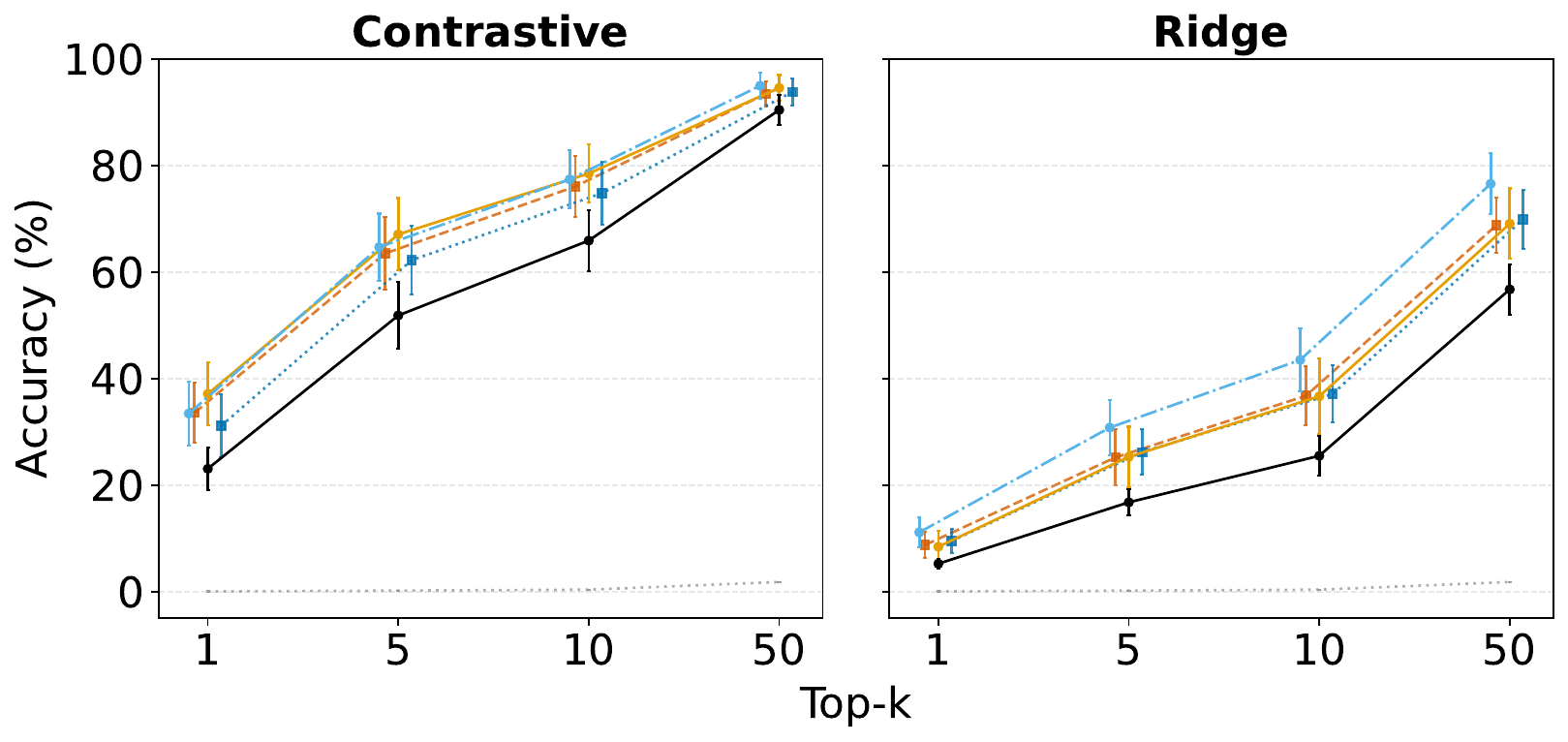}
    \caption{Top-k accuracy on NSD dataset}
    \label{fig:retrieval_nsd}
  \end{subfigure}\hfill
  \begin{subfigure}[t]{0.495\linewidth}
    \centering
    \includegraphics[width=\linewidth]{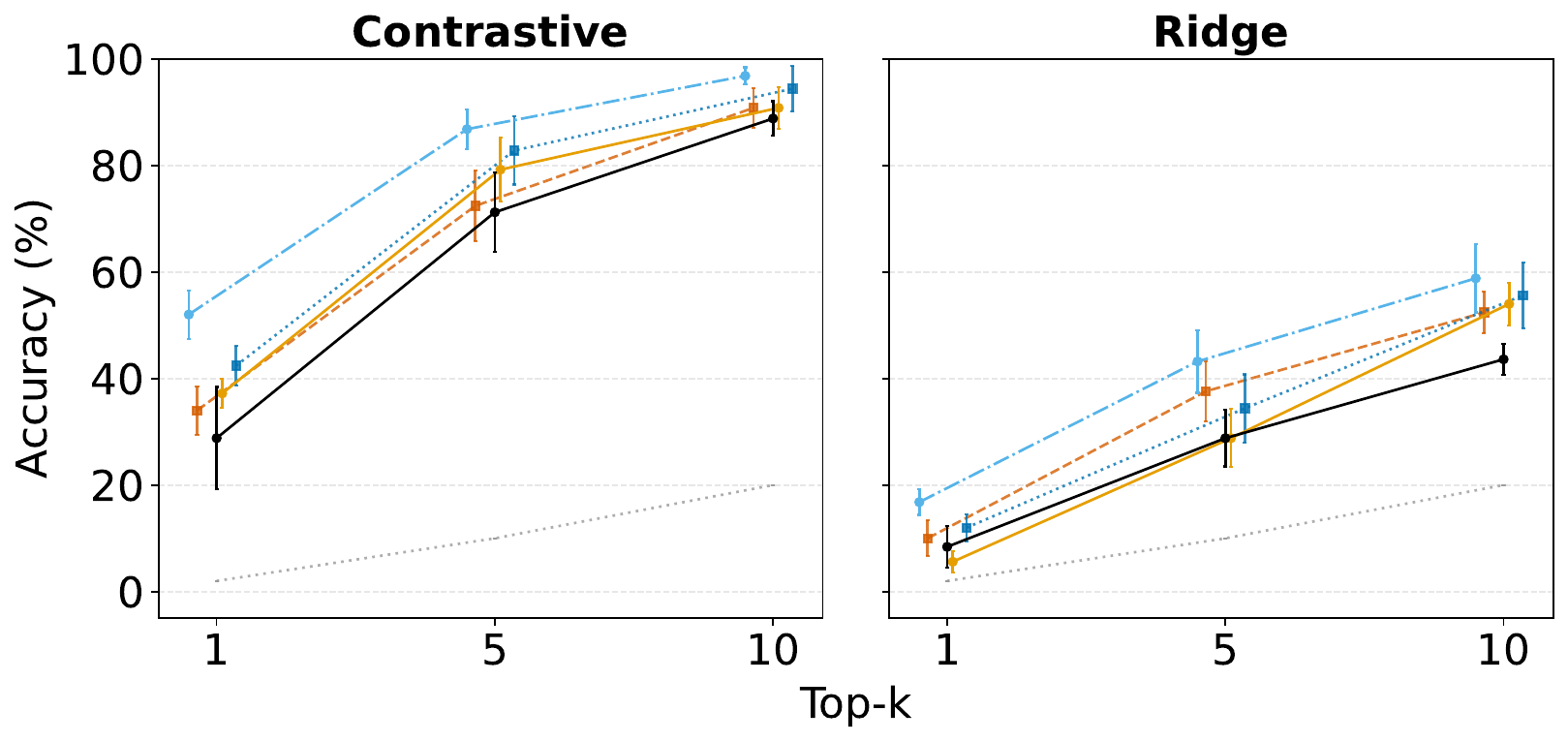}
    \caption{Top-k accuracy on GOD dataset}
    \label{fig:retrieval_god}
  \end{subfigure}

  \caption{fMRI-image retrieval accuracy of different models. Robust CLIP improves retrieval both on NSD and GOD. Error bars indicate variability across subjects.}
  \label{fig:combined}
\end{figure}

\textbf{Decoders with Adversarially Trained Models Show Stronger Alignment.}
Although the decoder is trained on a sample-wise objective, we observe consistent differences in global alignment across the dataset. Figure~\ref{alignment:a} summarizes both decoder-based alignment (Pearson correlation and cosine similarity) and representation-level alignment (RSA). Across all metrics, adversarially trained models (FARE and TeCoA) achieve higher alignment scores than standard CLIP, with TeCoA-4 consistently being the highest.

\vspace{3pt}\textbf{Decoders with Adversarially Trained Models Show Higher Retrieval Accuracy.}
We first re-implemented the contrastive decoding pipeline from ~\cite{efird2025finding}. Despite the differences in data split, we confirmed that the standard CLIP decoder is comparable to the original results. For fMRI-image retrieval, TeCoA and FARE decoders achieve consistently higher accuracy than the standard CLIP decoder, with the best-performing model (FARE-4) improving Top-1 accuracy by 13\% relative to the CLIP baseline. Figure \ref{fig:combined}-(a) shows this trend across 4 subjects and their mean. Under ridge regression---which yields lower overall performance than the contrastive decoding~\cite{efird2025finding}---the TeCoA and FARE decoders still outperform the CLIP decoder, while the best-performing model changed from FARE-4 to TeCoA-4. We demonstrate consistent improvements under both contrastive learning and ridge regression, suggesting that the choice of target representation plays a more significant role than the training objective. Detailed results are provided in \textbf{Appendix~\ref{sec:ap_alignment_retrieval}}, Table~\ref{tab:ap_summary_retrieval}.

\vspace{3pt}\textbf{Effect of Adversarial Strength on Decoding Performance.}
We further observe that, under the contrastive decoding setting, both FARE and TeCoA decoders achieve higher retrieval accuracy with strongly adversarially trained models (FARE-4 and TeCoA-4) compared to moderately robust models (FARE-2 and TeCoA-2). Interestingly, this trend does not consistently hold under ridge regression. In this setting, FARE-2 and FARE-4 decoders exhibit closely overlapping performance, whereas the TeCoA-4 decoder continues to outperform the TeCoA-2 decoder. 

\vspace{3pt}\textbf{TeCoA and FARE Exhibit Metric-Dependent Performance Difference.}
While both consistently outperform the standard CLIP decoder, we further observe that TeCoA and FARE exhibit different performance tendencies, such as relative ranking, depending on metrics. Specifically, TeCoA decoders tend to achieve higher scores on representation-level alignment metrics, including Pearson correlation, cosine similarity, and RSA. In contrast, FARE decoders achieve higher fMRI-image retrieval performance under the contrastive decoding setting. This indicates that no single adversarial training strategy uniformly dominates across all evaluation criteria. Instead, despite both being derived from adversarial training, TeCoA and FARE induce distinct representational properties that are emphasized differently depending on the metric.

\subsection{Decoders with Adversarially Trained Models Show Better Zero-Shot Capacity on the GOD Dataset}

\begin{figure}
    \vspace{-2pt}
    \centering
    \includegraphics[width=0.85\linewidth, height=0.2\textheight]{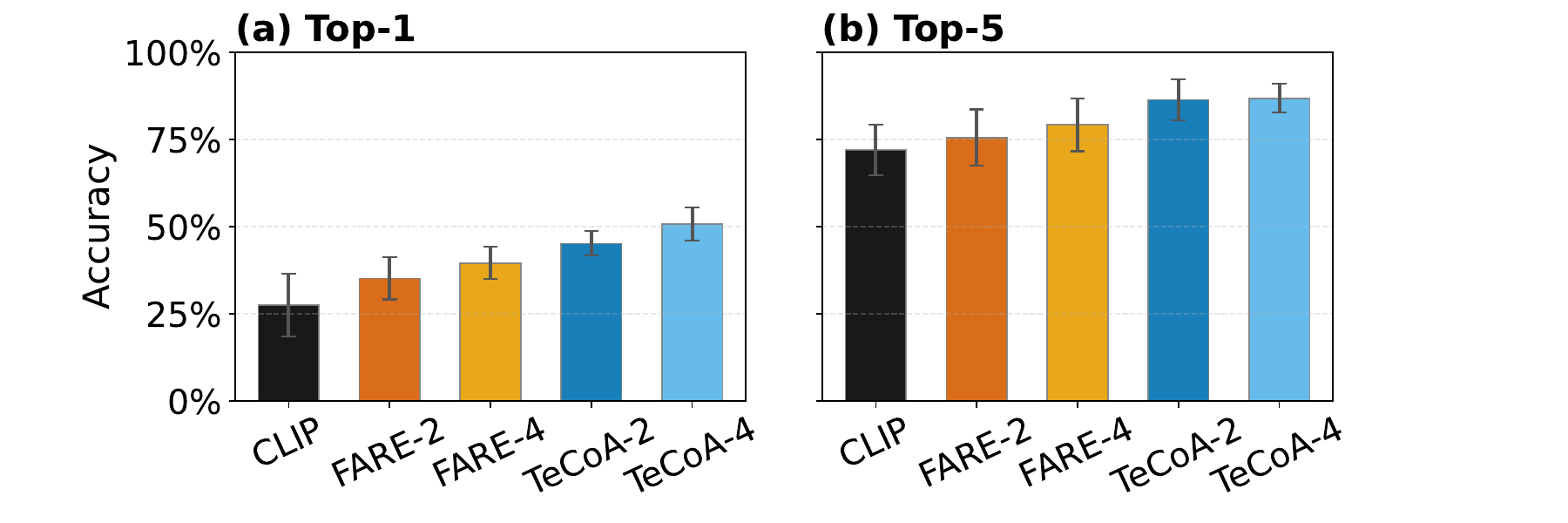}
    \caption{Top-k fMRI-text zero-shot classification on the GOD dataset. Error bars indicate variability across subjects.}
    \label{fig:zeroshot}
    \vspace{-2pt}
\end{figure}

Recent criticism~\cite{shirakawa2025spurious} has raised concerns that some CLIP-based pipelines~\cite{Li_2025neuraldiffuser,scotti2023reconstructingmindseyefmritoimage} evaluated on the NSD dataset may exhibit limited cross-dataset generalizability, indicating potential evaluation bias, such as double dipping~\cite{kriegeskorte2009circular}. In this section, we rely on the GOD dataset, which enforces zero-shot settings with non-overlapping train-test semantics, to investigate whether the observed improvements on the NSD dataset are reproducible and generalize under stricter semantic separation. 

\vspace{3pt}\textbf{Decoders with Adversarially Trained Models Maintain Generalizability in Zero-Shot fMRI-image retrieval.}
Figure~\ref{fig:combined}-(b) shows fMRI-image retrieval accuracy under contrastive decoding and ridge regression on the GOD dataset. Consistent with the NSD results, decoders with adversarially trained variants show higher performance than the standard CLIP decoder. In terms of zero-shot accuracy, the contrastive decoding setting shows TeCoA decoders consistently outperform FARE decoders, with TeCoA-4 being the best-performing model, while ridge regression shows this trend clearly only between strongly adversarially trained models (TeCoA-4 and FARE-4). Notably, FARE-4 exhibits low Top-1 accuracy, but recovers at higher Top-K values. 

\vspace{3pt}\textbf{Decoders with Adversarially Trained Models also Show Higher Zero-Shot fMRI-Text Classification Performance.}
We further evaluate text-prompt-based zero-shot classification~\cite{radford2021learningtransferablevisualmodels} using each model's text encoder. Following standard practice in machine learning ~\cite{radford2021learningtransferablevisualmodels, dhillon2020baselinefewshotimageclassification} and brain decoding literature~\cite{brainclip, yu2025metalearningincontexttransformermodel}, we construct prompts of the form ``a photo of \{cls\}'' and encode them to text embeddings, which are subsequently compared with predicted vision embeddings from decoders via cosine similarity for classification. 
This yields an fMRI-text zero-shot classifier. Consistent with retrieval results, both TeCoA and FARE decoders outperform the standard CLIP decoder, while TeCoA decoders also consistently outperform FARE decoders. The results are summarized in Figure~\ref{fig:zeroshot}.

\begin{figure}
  \centering
  \begin{subfigure}{0.49\linewidth}
    \centering
    \includegraphics[height=0.25\textheight,width=\linewidth]{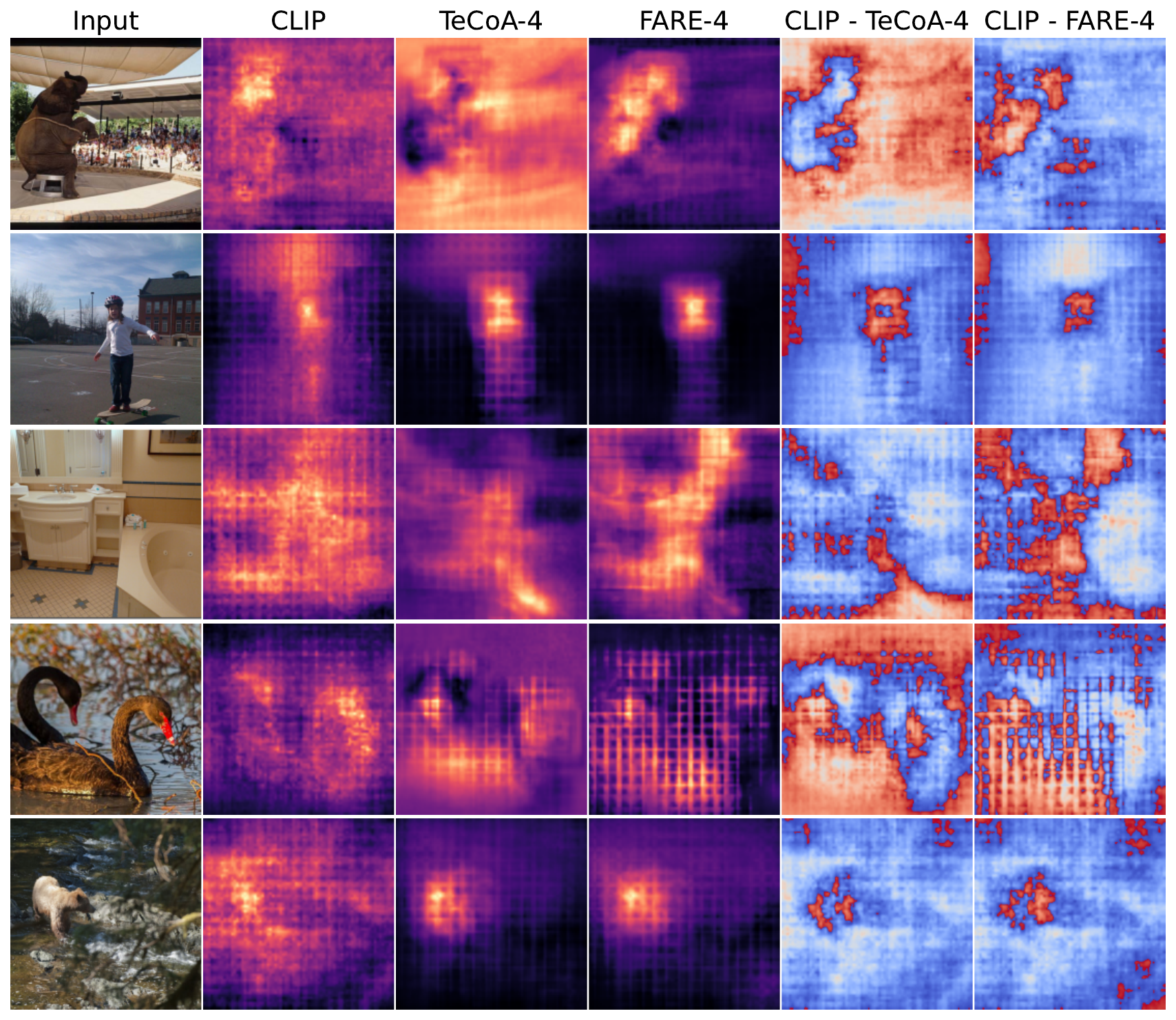}
  \end{subfigure}
  \hspace{0.0001\linewidth}
  \begin{subfigure}{0.49\linewidth}
  \centering
    \includegraphics[height=0.25\textheight,width=\linewidth]{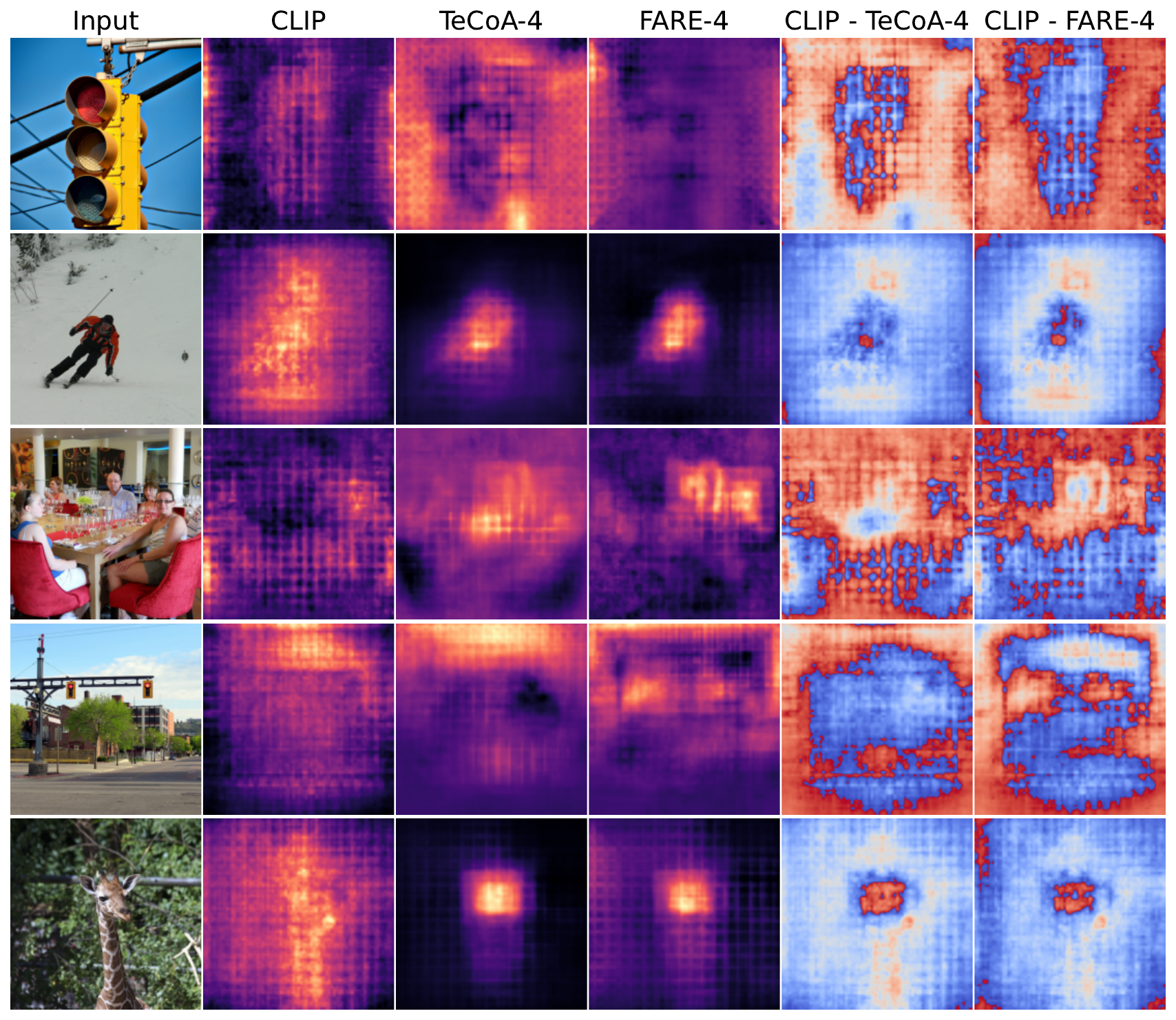}
  \end{subfigure}
  \caption{Attribution maps of 10 randomly selected images from the test dataset. From left to right: input image, CLIP, TeCoA-4, FARE-4, difference between CLIP and TeCoA-4, and difference between CLIP and FARE-4. All the attribution maps are generated from Subject 01.}
    \label{fig:att_map}
\end{figure}

\section{Attribution Analysis}

Despite the observed improvements in neural decoding performance, it remains unclear why decoders targeting adversarially robust representations exhibit such improvement and how this performance increase connects to the specific image features that decoders may rely on differently. To investigate this, we conduct an attribution-based analysis of feature utilization across decoders, following the framework in~\cite{palazzo2020decoding}. Our findings are twofold: (1) decoders trained on adversarially robust CLIP variants exhibit less spatially diffuse and more concentrated attribution patterns than the standard CLIP decoder; (2) these differences extend beyond simple refinement, with robust representations inducing systematic shifts in feature importance, leading to structurally distinct attribution patterns.

\subsection{Methodology}
We adopt the attribution framework of Palazzo et al.~\cite{palazzo2020decoding}, which estimates pixel-level contributions to neural–visual compatibility without requiring an additional generative model. Full methodological details are provided in the \textbf{Appendix~\ref{sec:ap_attr_method}}. 
We adapt the original framework by replacing the neural encoder with the trained fMRI decoder. Specifically, given fMRI data $x$ and an image $v$, we define the compatibility score as:
\begin{equation}
    F(x,v) = d(x)^\top \phi(v),
\end{equation}
where $d(\cdot)$ is the decoder and $\phi(\cdot)$ is the vision encoder (CLIP, TeCoA, or FARE).

Pixel-wise attribution is computed via occlusion by masking local regions of the image and measuring the change in compatibility:
\begin{equation}
    S(p,x,v) = F(x,v) - F(x, m(p)\odot v),
\end{equation}
where $m(p)$ masks a patch centered at pixel $p=(i,j)$. We apply this procedure across multiple scales and aggregate the results.

Both $d(x)$ and $\phi(v)$ are $L_2$-normalized, making $F(x,v)$ equivalent to cosine similarity. The resulting attribution map reflects the contribution of each image region to the alignment between fMRI responses and the target representation. We interpret these maps as \textbf{importance maps}, indicating which visual features each decoder relies on. Example attribution maps are shown in Figure~\ref{fig:att_map}.

\subsection{Results}
\textbf{Both Standard CLIP and Robust Variants Exhibit High Global Spatial Diffusion.}
To quantify the spatial distribution of attribution, we convert each importance map into a probability distribution over pixels and compute entropy over the flattened spatial domain. Let $s_i$ denote the attribution score at pixel $i$, and define
\begin{equation}
   p_i = \frac{s_i}{\sum_j s_j}.
\end{equation}

We measure entropy as:
\begin{equation}
   H(p) = - \sum_i p_i \log p_i.
\end{equation}

To enable comparison across images, we normalize entropy by its maximum value $\log(HW)$ where $H$ and $W$ denote the width and height of the attribution map:
\begin{equation}
   \tilde{H}(p) = \frac{H(p)}{\log(HW)} \in [0,1].
\end{equation}

We observe that all models, including CLIP and adversarially robust variants (TeCoA, FARE), exhibit consistently high normalized entropy values, indicating broadly distributed spatial attribution across pixels (see \textbf{Appendix~\ref{sec:at_aggregaion}}, Table~\ref{tab:ap_entropy_gini}). This suggests that global diffusion alone does not distinguish the models.

\vspace{3pt}\textbf{Robust Variants Exhibit Stronger Scale-Dependent Structural Variation in Attribution Maps.}
While global entropy is saturated, it does not capture differences in spatial structure across scales. To analyze this, we compute multi-scale attribution maps by applying max pooling with varying strides, producing coarse-grained versions of the original importance maps. Entropy is then computed for each scale.

\begin{figure}
  \centering
  \includegraphics[width=\linewidth]{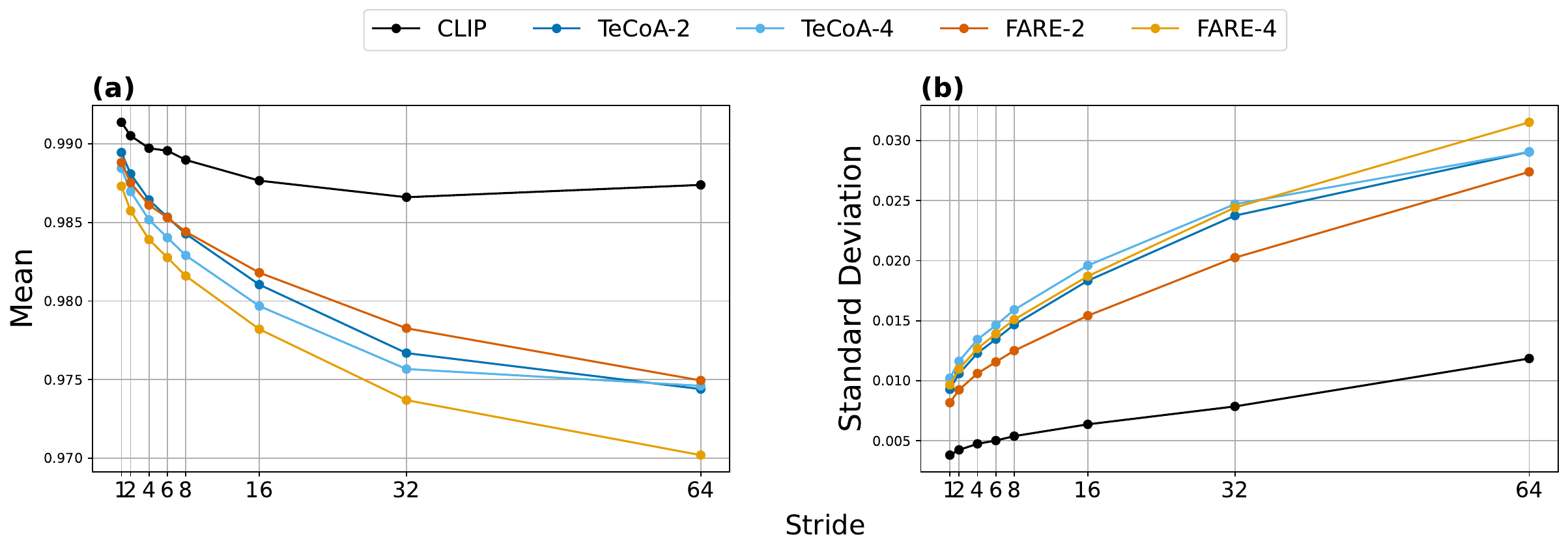}
  \caption{(a) Mean of Normalized Entropy measured by stride. (b) Standard Deviation of Normalized Entropy measured by stride.}
  \label{fig:entropy}
\end{figure}

Figure~\ref{fig:entropy}-(a) shows the mean entropy across scales. We observe that CLIP maintains relatively stable entropy across spatial resolutions, indicating limited sensitivity to changes in spatial scale. In contrast, both TeCoA and FARE exhibit a consistent decrease in entropy as spatial resolution becomes coarser, suggesting more structured and scale-dependent attribution patterns. We further examine variability across samples in Figure~\ref{fig:entropy}-(b). CLIP shows relatively stable entropy variance across scales, whereas robust models exhibit larger changes in variance under coarse-graining, indicating greater variability in spatial attribution patterns across images. We observe similar trends when using the Gini index~\cite{hurley2009comparingmeasuressparsity}, as reported in the \textbf{Appendix~\ref{sec:at_att_maps}}, Figure~\ref{fig:entropy_gini_combined}-(a).

Overall, these results suggest that while global diffusion is similar across models, robust variants exhibit stronger scale-dependent structure in attribution behavior.

\vspace{3pt}\textbf{Decoders with Adversarially Trained Models Rely on Spatially Distinct Image Features from Standard CLIP.}
As a follow-up analysis, we examine whether decoders trained with FARE and TeCoA rely on the same image features as the CLIP-based decoder. To this end, we quantify both spatial overlap and rank consistency between attribution maps using top-$k$ Intersection-over-Union (IoU) and Spearman rank correlation, respectively.

\begin{figure}
    \vspace{-2pt}
    \centering
    \includegraphics[width=0.75\linewidth]{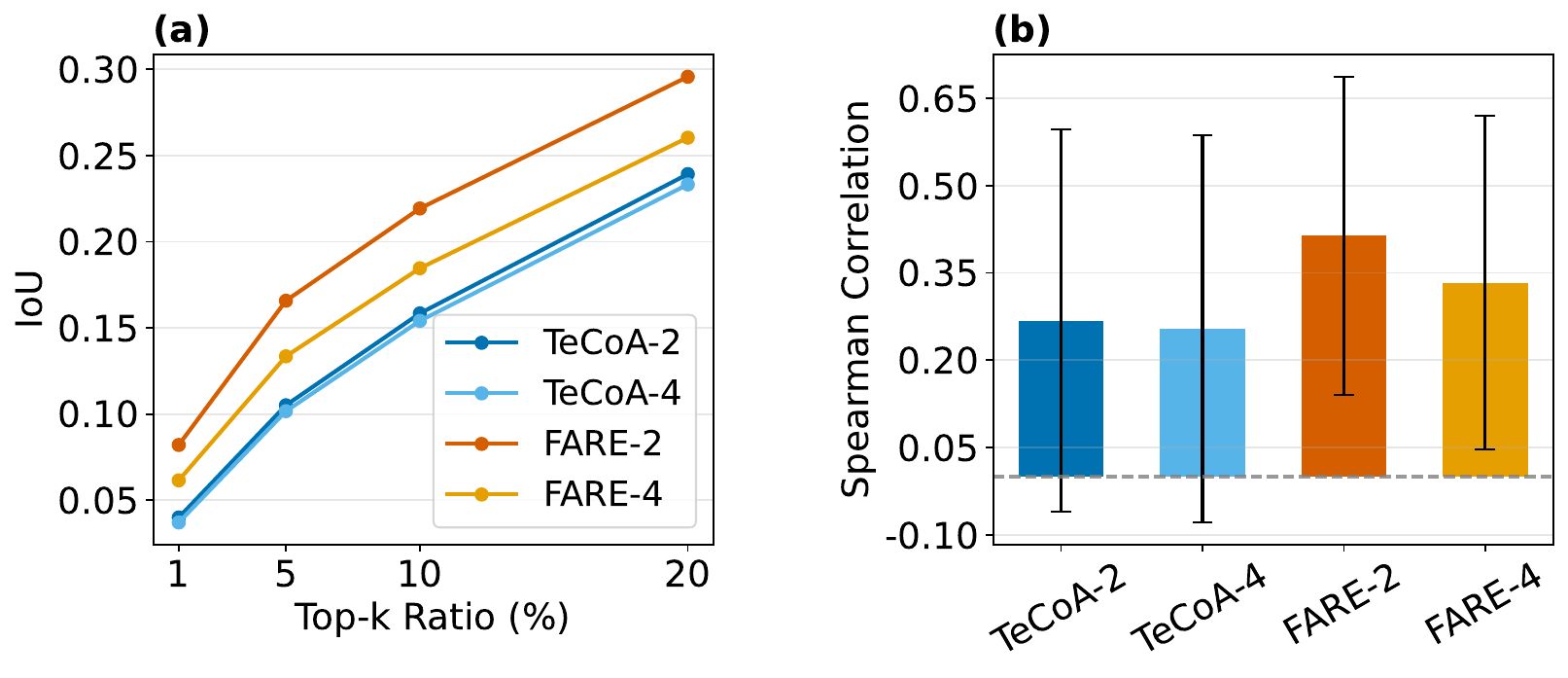}
    \caption{(a) Average Top-k IoU with CLIP (b) Average Spearman Correlation with CLIP. Both plots are generated from attribution maps. The plots are made from Subject 1, while the same trend is observed in subjects 1, 2, 5, and 7 (See \textbf{Appendix\ref{sec:at_aggregaion}}, Table~\ref{tab:at_spearman} and Table~\ref{tab:average_topk}).}
    \label{fig:avg_spr}
    \vspace{-2pt}
\end{figure}

Figure~\ref{fig:avg_spr} summarizes the results. We observe consistently low IoU across all top-$k$ thresholds, including at larger regions (e.g., 20\%), indicating that the spatial locations of highly attributed pixels differ substantially between CLIP and robust variants. Notably, IoU at the top-1\% level remains extremely low (below 0.1 for all models), suggesting minimal overlap in the most salient regions.

Spearman rank correlation further reveals moderate but highly variable agreement across images. While some samples exhibit high correlation, many show weak or even negative correlation, indicating substantial reordering of feature importance. Together, these results suggest that adversarially trained models do not simply refine CLIP’s feature usage, but instead rely on structurally distinct attribution patterns for alignment.

Interestingly, TeCoA and FARE exhibit different ranges of IoU and correlation values, suggesting that different adversarial training strategies may induce distinct modes of feature reorganization.

\vspace{-2pt}
\section{Discussion}
\vspace{-2pt}
\textbf{Interpretation of Attribution Patterns and Spurious Features.}
While we observe that adversarially robust models improve brain decoding performance and exhibit spatially distinct attribution structures from the standard CLIP decoder, the interpretability of these differences remains limited. In particular, attribution maps do not provide ground-truth correspondence to human visual saliency, making it difficult to determine whether highlighted regions reflect semantically meaningful features for human vision or model-specific predictive cues.
We additionally observe qualitative cases where the standard CLIP decoder seems to assign non-negligible importance to background regions (see \textbf{Appendix~\ref{sec:at_att_maps}}, Figure~\ref{fig:at_att_map_jsd}). While such behavior has been associated with shortcut-like feature usage in vision-language models~\cite{wang2024soberlookrobustnessclips, li2024closerlookexplainabilitycontrastive}, it is unclear whether these cues are irrelevant for neural processing, as human perception may also leverage contextual information. In addition, with our current experiment setting, we were not able to find clear statistical evidence that the CLIP decoder assigns more to background information due to the multi-context nature of NSD image stimuli. We therefore interpret these results as differences in representational usage rather than definitive evidence of spurious feature reliance.

\vspace{-2pt}
\section{Conclusion} 
\vspace{-2pt}
In this work, we investigated whether adversarial robustness in CLIP representations improves neural decoding performance and alignment with fMRI signals. Across two datasets (NSD and GOD), we found that decoders targeting adversarially trained representations consistently outperform the standard CLIP decoder in fMRI-image retrieval and zero-shot classification, while also achieving higher alignment scores measured by Pearson correlation, cosine similarity, and RSA.
Beyond performance gains, our attribution analysis reveals that adversarially trained representations induce structurally distinct feature usage patterns compared to standard CLIP, including differences in spatial concentration and scale-dependent attribution behavior. These findings suggest that robustness not only affects accuracy but also reorganizes how visual features are utilized in brain decoding.
Overall, our results highlight that the choice of target representation is a critical factor in neural decoding pipelines, and that adversarial robustness provides a meaningful inductive bias for improving alignment between computational models and brain activity.

\textbf{Acknowledgements}
This work was partially supported by JSPS KAKENHI Grant JP24H00732, by JST CREST Grants JPMJCR20D3 and JPMJCR2562 including AIP challenge program, and by JST K Program Grant JPMJKP24C2 Japan


\medskip
\bibliographystyle{plain}
\bibliography{references}

\appendix

\section{Experiment Details}
\subsection{Training Scheme}
\paragraph{Implementation Details.}
\label{sec:ap_implementation}

We provide additional implementation details for training the linear brain decoder. 
Our training protocol follows prior work~\cite{efird2025finding}. Specifically, we train a linear contrastive decoder for 5000 iterations. Data augmentation is applied to the fMRI inputs by adding normalized Gaussian noise scaled by a factor of $0.1$. We use a learning rate of $1 \times 10^{-3}$, a temperature parameter of $0.03$, and a batch size of $128$. Training is performed without a learning rate scheduler. The same hyperparameter configuration is used for all GOD decoders.

For ridge regression, we perform a grid search over the regularization parameter $\lambda$ and select the model with the best validation performance.

\vspace{-0.5em}

\paragraph{Adversarial Training Objectives.}
TeCoA~\cite{mao2022understanding} introduces a text-supervised adversarial fine-tuning objective. 
Let $\{(x_i, y_i)\}_{i=1}^{n}$ denote the training dataset. The objective is defined as:
\begin{equation}
    L_{\mathrm{TeCoA}}(y, f(\phi, x)) 
    = -\log \left( 
    \frac{e^{f_{y}(\phi, x)}}{\sum_{k=1}^{K} e^{f_k(\phi, x)}} 
    \right),
\end{equation}
where $f(\phi, x)$ denotes the classifier logits produced by the model.

While text supervision has been reported to degrade CLIP's zero-shot performance, Schlarmann et al.~\cite{schlarmann2024robust} proposed FARE, a self-supervised adversarial fine-tuning method that preserves zero-shot capability. 
Let $\phi_{\mathrm{Org}}$ denote the original CLIP encoder. The FARE objective is given by:
\begin{equation}
    L_{\mathrm{FARE}}(\phi, x) 
    = \max_{\|z - x\|_{\infty} \leq \varepsilon} 
    \|\phi(z) - \phi_{\mathrm{Org}}(x)\|_2^2.
\end{equation}

\vspace{-0.5em}
\paragraph{Computing Resources}

All linear decoder training experiments were conducted on a single \mbox{NVIDIA RTX 6000 Ada Generation} GPU. For the NSD dataset, a single training run required approximately 10 minutes, while training on the GOD dataset completed in under one minute. 

Attribution map generation was performed on a high-performance computing cluster using a single NVIDIA H100 SXM5 GPU (94GB HBM). This process required approximately 4--5 days per subject for 982 images.

\subsection{Datasets}
\label{sec:ap_dataset}

In this section, we further report the details of the dataset settings 
\vspace{-0.3em}
\subsubsection{NSD Dataset}

In the NSD dataset, each subject has 1--3 times repeated betas on each image, depending on original experiment completion and held-out data for the Algonaut project~\cite{gifford2023algonautsproject2023challenge}. 
We followed the fMRI preprocessing settings of the previous work~\cite{efird2025finding}, which selected denoised beta voxels (\textit{betas\_fithrf\_GLMdenoise\_RR}) by thresholding based on noise ceiling. To avoid possible double dipping~\cite{kriegeskorte2009circular}, this noise ceiling was computed using only the train dataset split in advance. Although ~\cite{allen2022massive} introduced a data split to make sure 3 repeated betas for the test dataset, we arranged the test set back to the original NSD test set~\cite{allen2022massive, brainclip}, which contains subject-shared 982 image stimuli and corresponding fMRI betas trials to test the consistency of the results on the same stimuli. For this setting, subjects 1, 2, 5, and 7 have full fMRI trials (3 repeated).

\paragraph{Preprocessing Parameters.} In our experiment, voxel selection was done by threshold $8.0$. After voxel selection, we did not apply further parcellations, such as via Region of Interest (ROI). Lastly, we normalized the selected voxels by session. This procedure and the parameter values followed previous work's setting~\cite{efird2025finding}. 
\vspace{-0.3em}
\subsubsection{GOD Dataset}

In the GOD dataset, test images are semantically disjoint from training images, with 100 categories used for training and 50 for testing. We use the preprocessed beta signals provided by the dataset authors, which are parcellated into Regions of Interest (ROI). In our experiments, we use all voxels within the visual cortex without additional preprocessing or feature selection.

\subsection{Neural Alignment and fMRI-image Retrieval.}

\label{sec:ap_alignment_retrieval}

\begin{table}[]
    \centering
    \caption{Summary of the Representational Similarity Analysis (RSA) results.  }
    \label{tab:at_rsa}
    \vspace{0.5em}
    \begin{tabular}{lrrrrr}
    \toprule
     & RSA (Subj01) & RSA (Subj02) & RSA (Subj05) & RSA (Subj07) & Mean \\
    \midrule
    CLIP & 0.279 & 0.283 & 0.308 & 0.269 & 0.285 \\
    FARE-2 & 0.348 & 0.354 & 0.371 & 0.327 & 0.350 \\
    FARE-4 & 0.365 & 0.369 & 0.388 & 0.344 & 0.366 \\
    TeCoA-2 & 0.392 & 0.382 & 0.405 & 0.364 & 0.386 \\
    TeCoA-4 & 0.433 & 0.428 & 0.444 & 0.397 & 0.425 \\
    \bottomrule
    \end{tabular}
\end{table}

\begin{table}[]
    \centering
    \caption{Summary of the Centered Kernel Alignment (CKA) results.}
    \label{tab:at_cka}
    \vspace{0.5em}
    \begin{tabular}{lrrrrr}
    \toprule
     & CKA (Subj01) & CKA (Subj02) & CKA (Subj05) & CKA (Subj07) & Mean \\
    \midrule
    CLIP & 0.408 & 0.416 & 0.448 & 0.393 & 0.416 \\
    FARE-2 & 0.446 & 0.460 & 0.478 & 0.420 & 0.451 \\
    FARE-4 & 0.461 & 0.472 & 0.493 & 0.432 & 0.464 \\
    TeCoA-2 & 0.461 & 0.460 & 0.484 & 0.427 & 0.458 \\
    TeCoA-4 & 0.489 & 0.492 & 0.508 & 0.444 & 0.483 \\
    \bottomrule
    \end{tabular}
\end{table}
\paragraph{Neural Alignment between Models and fMRI Responses.} 
Pearson correlation and cosine similarity were measured between the latent embeddings of models and predicted embeddings from trained decoders, while Representation Similarity Analysis (RSA) was computed from the two Representation Difference Matrices (RDMs) between latent embeddings of models and fMRI responses, directly. All the metrics are measured subject-wise on the test dataset to observe inter-subject variability. We additionally report the Centered Kernel Alignment (CKA) results as well in Table~\ref{tab:at_cka}, which was measured between fMRI responses and latent embeddings of models on the test dataset, following the same procedure as RSA.

\paragraph{fMRI-image Retrieval.}
\label{sec:ap:fMRI-retrieval}
We present the comprehensive experimental results for fMRI-image retrieval. The Table~\ref{tab:ap_summary_retrieval} shows summary results of Top-K accuracy with mean and variation, and Table~\ref{tab:ap_subj_retrieval} shows Subject-Wise Top-1 ACC on the NSD dataset. 
\FloatBarrier

\begin{table}[]
    \centering
    \caption{Summary results of fMRI-image retrieval by Top-K Accuracy on the GOD and NSD datasets. }
    \label{tab:ap_summary_retrieval}
    \vspace{0.5em}
    \begin{tabular}{lllllll}
    \toprule
     &  & k & 1 & 5 & 10 & 50 \\
    Dataset & Model & Method &  &  &  &  \\
    \midrule
    \multirow[t]{10}{*}{GOD} & \multirow[t]{2}{*}{CLIP} & Contrastive & 0.29 $\pm$ 0.11 & 0.71 $\pm$ 0.08 & 0.89 $\pm$ 0.04 & -- \\
     &  & Ridge & 0.08 $\pm$ 0.04 & 0.29 $\pm$ 0.06 & 0.44 $\pm$ 0.03 & -- \\
    \cline{2-7}
     & \multirow[t]{2}{*}{FARE-2} & Contrastive & 0.34 $\pm$ 0.05 & 0.72 $\pm$ 0.07 & 0.91 $\pm$ 0.04 & -- \\
     &  & Ridge & 0.10 $\pm$ 0.04 & 0.38 $\pm$ 0.06 & 0.52 $\pm$ 0.04 & -- \\
    \cline{2-7}
     & \multirow[t]{2}{*}{FARE-4} & Contrastive & 0.37 $\pm$ 0.03 & 0.79 $\pm$ 0.07 & 0.91 $\pm$ 0.04 & -- \\
     &  & Ridge & 0.06 $\pm$ 0.02 & 0.29 $\pm$ 0.06 & 0.54 $\pm$ 0.04 & -- \\
    \cline{2-7}
     & \multirow[t]{2}{*}{TeCoA-2} & Contrastive & 0.42 $\pm$ 0.04 & 0.83 $\pm$ 0.07 & 0.94 $\pm$ 0.05 & -- \\
     &  & Ridge & 0.12 $\pm$ 0.03 & 0.34 $\pm$ 0.07 & 0.56 $\pm$ 0.07 & -- \\
    \cline{2-7}
     & \multirow[t]{2}{*}{TeCoA-4} & Contrastive & 0.52 $\pm$ 0.05 & 0.87 $\pm$ 0.04 & 0.97 $\pm$ 0.02 & -- \\
     &  & Ridge & 0.17 $\pm$ 0.03 & 0.43 $\pm$ 0.07 & 0.59 $\pm$ 0.07 & -- \\
    \cline{1-7} \cline{2-7}
    \multirow[t]{10}{*}{NSD} & \multirow[t]{2}{*}{CLIP} & Contrastive & 0.20 $\pm$ 0.05 & 0.46 $\pm$ 0.09 & 0.60 $\pm$ 0.09 & 0.88 $\pm$ 0.05 \\
     &  & Ridge & 0.04 $\pm$ 0.01 & 0.15 $\pm$ 0.04 & 0.23 $\pm$ 0.05 & 0.53 $\pm$ 0.07 \\
    \cline{2-7}
     & \multirow[t]{2}{*}{FARE-2} & Contrastive & 0.30 $\pm$ 0.07 & 0.58 $\pm$ 0.09 & 0.71 $\pm$ 0.09 & 0.92 $\pm$ 0.04 \\
     &  & Ridge & 0.07 $\pm$ 0.03 & 0.22 $\pm$ 0.06 & 0.33 $\pm$ 0.07 & 0.65 $\pm$ 0.07 \\
    \cline{2-7}
     & \multirow[t]{2}{*}{FARE-4} & Contrastive & 0.33 $\pm$ 0.08 & 0.62 $\pm$ 0.10 & 0.74 $\pm$ 0.08 & 0.93 $\pm$ 0.04 \\
     &  & Ridge & 0.07 $\pm$ 0.03 & 0.22 $\pm$ 0.06 & 0.33 $\pm$ 0.08 & 0.66 $\pm$ 0.08 \\
    \cline{2-7}
     & \multirow[t]{2}{*}{TeCoA-2} & Contrastive & 0.27 $\pm$ 0.07 & 0.57 $\pm$ 0.09 & 0.70 $\pm$ 0.08 & 0.92 $\pm$ 0.05 \\
     &  & Ridge & 0.08 $\pm$ 0.03 & 0.23 $\pm$ 0.05 & 0.34 $\pm$ 0.07 & 0.66 $\pm$ 0.07 \\
    \cline{2-7}
     & \multirow[t]{2}{*}{TeCoA-4} & Contrastive & 0.30 $\pm$ 0.08 & 0.60 $\pm$ 0.09 & 0.73 $\pm$ 0.08 & 0.93 $\pm$ 0.04 \\
     &  & Ridge & 0.10 $\pm$ 0.03 & 0.28 $\pm$ 0.06 & 0.40 $\pm$ 0.07 & 0.73 $\pm$ 0.07 \\
    \cline{1-7} \cline{2-7}
    \bottomrule
    \end{tabular}
\end{table}

\begin{table}[]
    \centering
    \caption{Subject-wise Top-1 accuracy on the NSD dataset. 
    Across subject, decoders with adversarially trained CLIP variants consistently improve Top-1 accuracy relative to CLIP. }
    \label{tab:ap_subj_retrieval}
    \vspace{0.5em}
    \resizebox{\linewidth}{!}{
    \begin{tabular}{llrrrrrrrrr}
    \toprule
     & Subject & Subj01 & Subj02 & Subj03 & Subj04 & Subj05 & Subj06 & Subj07 & Subj08 & Mean \\
    Model & Method &  &  &  &  &  &  &  &  &  \\
    \midrule
    \multirow[t]{2}{*}{CLIP} & Contrastive & 0.25 & 0.21 & 0.16 & 0.17 & 0.28 & 0.22 & 0.18 & 0.12 & 0.20 \\
     & Ridge & 0.06 & 0.05 & 0.04 & 0.04 & 0.06 & 0.05 & 0.04 & 0.02 & 0.04 \\
    \cline{1-11}
    \multirow[t]{2}{*}{FARE-2} & Contrastive & 0.37 & 0.32 & 0.24 & 0.27 & 0.40 & 0.33 & 0.25 & 0.18 & 0.30 \\
     & Ridge & 0.09 & 0.08 & 0.05 & 0.06 & 0.12 & 0.08 & 0.06 & 0.04 & 0.07 \\
    \cline{1-11}
    \multirow[t]{2}{*}{FARE-4} & Contrastive & 0.41 & 0.36 & 0.28 & 0.29 & 0.43 & 0.36 & 0.28 & 0.20 & 0.33 \\
     & Ridge & 0.09 & 0.08 & 0.05 & 0.07 & 0.13 & 0.07 & 0.04 & 0.04 & 0.07 \\
    \cline{1-11}
    \multirow[t]{2}{*}{TeCoA-2} & Contrastive & 0.33 & 0.29 & 0.23 & 0.25 & 0.39 & 0.31 & 0.23 & 0.17 & 0.27 \\
     & Ridge & 0.09 & 0.09 & 0.07 & 0.07 & 0.13 & 0.10 & 0.07 & 0.05 & 0.08 \\
    \cline{1-11}
    \multirow[t]{2}{*}{TeCoA-4} & Contrastive & 0.36 & 0.32 & 0.26 & 0.27 & 0.41 & 0.34 & 0.24 & 0.17 & 0.30 \\
     & Ridge & 0.10 & 0.11 & 0.08 & 0.08 & 0.15 & 0.12 & 0.08 & 0.06 & 0.10 \\
    \cline{1-11}
    \bottomrule
    \end{tabular}
    }
\end{table}

\section{Attribution Maps}

\subsection{Methodology}
\label{sec:ap_attr_method}
Given neural representation $e$ and image $v$, compatibility is defined as:
\begin{equation}
    F(e,v) = \varphi(e)^T \theta(v),
\end{equation}
where $\varphi$ and $\theta$ are modality-specific encoders.

Pixel-wise attribution is computed via occlusion:
\begin{equation}
    S(p,\sigma,e,v) = F(e,v) - F(e, m_\sigma(p)\odot v),
\end{equation}
where $m_\sigma(p)$ masks a patch of size $\sigma$, centered at pixel coordinate $p=(i,j)$. The final attribution map aggregates across scales:
\begin{equation}
    S(p,e,v) = \sum_{\sigma} S(p,\sigma,e,v).
\end{equation}

We adapt this framework by replacing the neural encoder $\varphi(\cdot)$ with an fMRI decoder $d(\cdot)$ and the visual encoder $\theta$ with pretrained CLIP, TeCoA, and FARE. The compatibility score becomes:
\begin{equation}
    F(x,v) = d(x)^T \phi(v),
\end{equation}
where $x$ denotes fMRI voxel data and $\phi$ denotes the vision encoder (CLIP, TeCoA, and FARE).

Pixel-wise attribution is defined analogously:
\begin{equation}
    S(p,\sigma,x,v) = F(x,v) - F(x, m_\sigma(p)\odot v),
\end{equation}
with final aggregation:
\begin{equation}
    S(p,x,v) = \sum_{\sigma} S(p,\sigma,x,v).
\end{equation}

We further apply $L_2$-normalization to embeddings from $d$ and $\phi$, making the compatibility equivalent to cosine similarity as noted in the main paper.

\paragraph{Hyperparameters.} 
For the scale parameter $\sigma$, we used $3$, $5$, $9$, $17$, $33$, and $65$. For each pixel location $p$, attribution scores $S(p,\sigma,x,v)$ computed across different scales were summed and then normalized to obtain the final attribution map, following the implementation details from prior work~\cite{palazzo2020decoding}. To reduce computational cost, we additionally applied a spatial stride of $3$ when iterating perturbation locations. Concretely, attribution scores were evaluated every three pixels along both spatial dimensions rather than exhaustively at every pixel location.

\subsection{Aggregation and Statistics}
\label{sec:at_aggregaion}
\begin{figure}[t]
    \centering

    \begin{subfigure}{\linewidth}
        \centering
        \includegraphics[width=\linewidth]{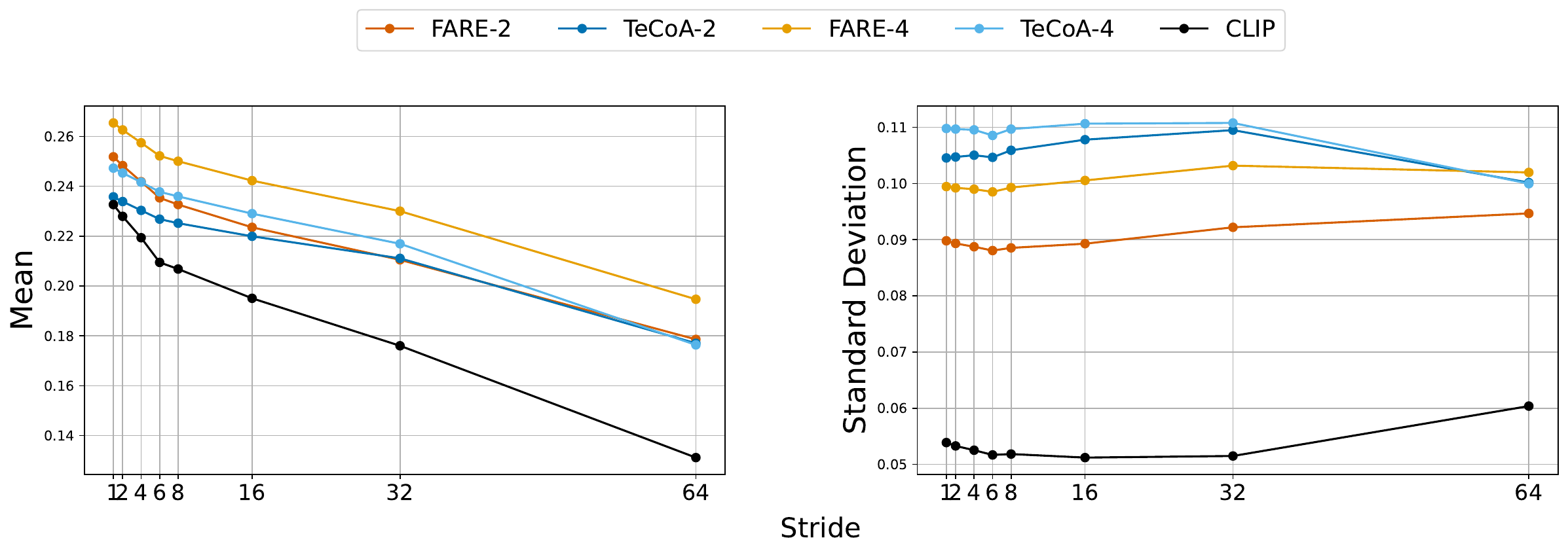}
        \caption{
        Mean and standard deviation of the Gini index across spatial scales. CLIP maintains relatively stable sparsity statistics across resolutions, whereas robust models exhibit larger changes under coarse-graining.
        }
        \label{fig:gini_mean_std}
    \end{subfigure}
    \vspace{0.5em}
    \begin{subfigure}{\linewidth}
        \centering
        \includegraphics[width=\linewidth]{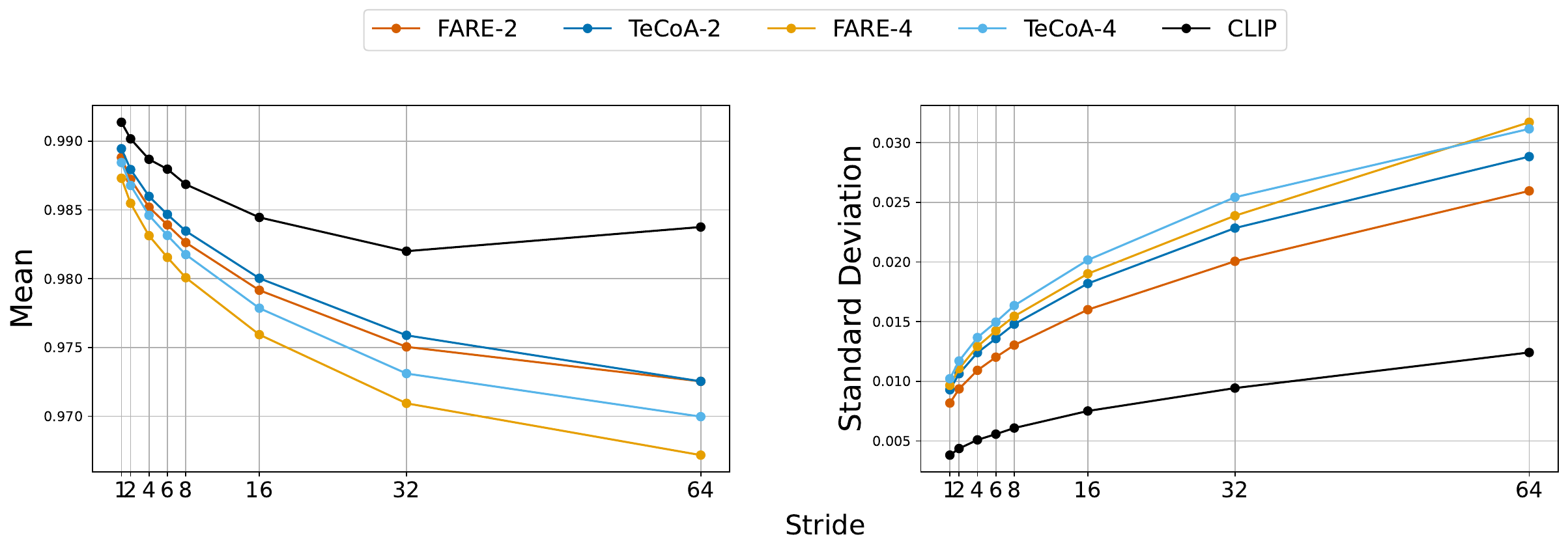}
        \caption{
        Scale-dependent entropy mean and standard deviation measured with average pooling. Similar to the original max-pooling implementation, consistent contrast is observed between CLIP and robust variants.
        }
        \label{fig:ap_entropy_mean_std_apool}
    \end{subfigure}

    \caption{
    Statistics of the scale-dependent Gini index~\cite{hurley2009comparingmeasuressparsity} and average-pooled entropy across models. Robust variants exhibit larger changes in attribution distributions under spatial coarse-graining compared to CLIP.
    }
    \label{fig:entropy_gini_combined}

\end{figure}

Here, we provide comprehensive measurement for the metrics introduced in the main paper for attribution maps, Entropy, Top-K IoU, and Spearman correlation. We additionally introduce the summary of Gini index~\cite{hurley2009comparingmeasuressparsity} in Figure~\ref{fig:entropy_gini_combined}-(a) as supplementary statistics of the scale-dependent sparsity measure. Gini index is broadly used for sparsity measure of the saliency maps~\cite{chalasani2020conciseexplanationsneuralnetworks, gong2025boosting}. We show that our results show consistent trends where coarse-grained statistics between CLIP and adversarially robust CLIP variants constantly diverge, indicating a structural difference in attribution patterns.

We also offer average pooling implemented version of scale-dependent entropy measured by stride in Figure~\ref{fig:entropy_gini_combined}-(b). As max-pool operation may induce bias in sampling of pixels, we alternatively employ average pooling to check consistency of the results. Although mean and standard deviation of the original CLIP show relative changes in magnitude, Figure~\ref{fig:entropy_gini_combined}-(b) shows the trend and contrast between CLIP and robust CLIP variants are consistent through strides. 

In terms of Top-K IoU and Spearman correlation, the results are summarized from Table~\ref{tab:average_topk} and Table~\ref{tab:at_spearman} respectively. The results are grouped by subject as our attribution maps were generated from these subjects for consistent comparison across shared image sets.  

\begin{table}[]
    \centering
    \caption{Average Entropy and Gini Index of the generated attribution maps per model. The results are grouped by subjects.}
    \label{tab:ap_entropy_gini}
    \vspace{0.5em}
    \begin{tabular}{llcccccc}
    \toprule
    & & \multicolumn{3}{c}{Entropy} & \multicolumn{3}{c}{Gini} \\
    \cmidrule(r){3-5} \cmidrule(r){6-8}
    Subject & Model & Mean & Min & Max & Mean & Min & Max \\
    \midrule
    \multirow{5}{*}{Subj01} & FARE-2 & 0.989 & 0.949 & 0.999 & 0.252 & 0.056 & 0.559 \\
     & TeCoA-2 & 0.989 & 0.944 & 0.999 & 0.236 & 0.052 & 0.589 \\
     & FARE-4 & 0.987 & 0.939 & 0.999 & 0.265 & 0.050 & 0.612 \\
     & TeCoA-4 & 0.988 & 0.934 & 0.999 & 0.247 & 0.050 & 0.628 \\
     & CLIP & 0.991 & 0.973 & 0.999 & 0.233 & 0.078 & 0.420 \\
    \midrule
    \multirow{5}{*}{Subj02} & FARE-2 & 0.989 & 0.948 & 0.999 & 0.246 & 0.051 & 0.564 \\
     & TeCoA-2 & 0.990 & 0.920 & 0.999 & 0.232 & 0.033 & 0.659 \\
     & FARE-4 & 0.988 & 0.931 & 0.999 & 0.259 & 0.056 & 0.627 \\
     & TeCoA-4 & 0.989 & 0.910 & 0.999 & 0.242 & 0.045 & 0.694 \\
     & CLIP & 0.991 & 0.968 & 0.999 & 0.235 & 0.084 & 0.460 \\
    \midrule
    \multirow{5}{*}{Subj05} & FARE-2 & 0.988 & 0.936 & 0.999 & 0.262 & 0.060 & 0.569 \\
     & TeCoA-2 & 0.989 & 0.941 & 1.000 & 0.241 & 0.039 & 0.599 \\
     & FARE-4 & 0.986 & 0.916 & 0.999 & 0.277 & 0.061 & 0.687 \\
     & TeCoA-4 & 0.988 & 0.921 & 1.000 & 0.255 & 0.052 & 0.660 \\
     & CLIP & 0.991 & 0.976 & 0.999 & 0.234 & 0.086 & 0.401 \\
    \midrule
    \multirow{5}{*}{Subj07} & FARE-2 & 0.989 & 0.943 & 0.999 & 0.248 & 0.049 & 0.590 \\
     & TeCoA-2 & 0.990 & 0.944 & 1.000 & 0.233 & 0.035 & 0.587 \\
     & FARE-4 & 0.988 & 0.931 & 0.999 & 0.262 & 0.056 & 0.632 \\
     & TeCoA-4 & 0.989 & 0.936 & 1.000 & 0.245 & 0.033 & 0.620 \\
     & CLIP & 0.991 & 0.971 & 0.999 & 0.232 & 0.063 & 0.431 \\
    \bottomrule
    \end{tabular}
\end{table}

\begin{table}[]
    \centering
    \caption{Average Spearman correlation between attribution maps from the standard CLIP and robust CLIP variants. Results are grouped by subject 1, 2, 5, and 7. Variability among samples are shown together.}
    \vspace{0.5em}
    \begin{tabular}{llc}
    \toprule
    Subject & Model & Spearman Correlation \\
    \midrule
    \multirow{4}{*}{Subj01} & TeCoA-2 & 0.268 $\pm$ 0.329 \\
     & TeCoA-4 & 0.254 $\pm$ 0.333 \\
     & FARE-2 & 0.414 $\pm$ 0.273 \\
     & FARE-4 & 0.333 $\pm$ 0.287 \\
    \midrule
    \multirow{4}{*}{Subj02} & TeCoA-2 & 0.267 $\pm$ 0.338 \\
     & TeCoA-4 & 0.257 $\pm$ 0.338 \\
     & FARE-2 & 0.416 $\pm$ 0.277 \\
     & FARE-4 & 0.335 $\pm$ 0.296 \\
    \midrule
    \multirow{4}{*}{Subj05} & TeCoA-2 & 0.284 $\pm$ 0.326 \\
     & TeCoA-4 & 0.268 $\pm$ 0.326 \\
     & FARE-2 & 0.429 $\pm$ 0.275 \\
     & FARE-4 & 0.350 $\pm$ 0.287 \\
    \midrule
    \multirow{4}{*}{Subj07} & TeCoA-2 & 0.273 $\pm$ 0.326 \\
     & TeCoA-4 & 0.258 $\pm$ 0.335 \\
     & FARE-2 & 0.419 $\pm$ 0.273 \\
     & FARE-4 & 0.330 $\pm$ 0.294 \\
    \bottomrule
    \end{tabular}
    
    \label{tab:at_spearman}
\end{table}

\begin{table}[]
    \centering  
    \caption{Average Top-K IoU between attribution maps from standard CLIP and robust CLIP variants. The results are grouped by subjects. Variability among samples are shown together.}
    \label{tab:average_topk}
    \vspace{0.5em}
    \begin{tabular}{llcccc}
    \toprule
    Subject & Model & Top-1\% & Top-5\% & Top-10\% & Top-20\% \\
    \midrule
    \multirow{4}{*}{Subj01} & TeCoA-2 & 0.040 $\pm$ 0.084 & 0.105 $\pm$ 0.126 & 0.158 $\pm$ 0.147 & 0.239 $\pm$ 0.157 \\
     & TeCoA-4 & 0.037 $\pm$ 0.082 & 0.102 $\pm$ 0.125 & 0.154 $\pm$ 0.145 & 0.233 $\pm$ 0.156 \\
     & FARE-2 & 0.082 $\pm$ 0.127 & 0.166 $\pm$ 0.160 & 0.219 $\pm$ 0.159 & 0.296 $\pm$ 0.149 \\
     & FARE-4 & 0.062 $\pm$ 0.110 & 0.133 $\pm$ 0.145 & 0.185 $\pm$ 0.149 & 0.260 $\pm$ 0.145 \\
    \midrule
    \multirow{4}{*}{Subj02} & TeCoA-2 & 0.038 $\pm$ 0.083 & 0.105 $\pm$ 0.125 & 0.161 $\pm$ 0.146 & 0.241 $\pm$ 0.159 \\
     & TeCoA-4 & 0.034 $\pm$ 0.075 & 0.100 $\pm$ 0.126 & 0.156 $\pm$ 0.147 & 0.236 $\pm$ 0.161 \\
     & FARE-2 & 0.080 $\pm$ 0.120 & 0.166 $\pm$ 0.152 & 0.221 $\pm$ 0.154 & 0.301 $\pm$ 0.153 \\
     & FARE-4 & 0.057 $\pm$ 0.103 & 0.133 $\pm$ 0.142 & 0.188 $\pm$ 0.148 & 0.266 $\pm$ 0.150 \\
    \midrule
    \multirow{4}{*}{Subj05} & TeCoA-2 & 0.043 $\pm$ 0.088 & 0.112 $\pm$ 0.131 & 0.165 $\pm$ 0.146 & 0.248 $\pm$ 0.157 \\
     & TeCoA-4 & 0.040 $\pm$ 0.084 & 0.108 $\pm$ 0.130 & 0.161 $\pm$ 0.145 & 0.243 $\pm$ 0.156 \\
     & FARE-2 & 0.094 $\pm$ 0.140 & 0.179 $\pm$ 0.164 & 0.233 $\pm$ 0.159 & 0.310 $\pm$ 0.153 \\
     & FARE-4 & 0.071 $\pm$ 0.122 & 0.147 $\pm$ 0.155 & 0.200 $\pm$ 0.156 & 0.278 $\pm$ 0.153 \\
    \midrule
    \multirow{4}{*}{Subj07} & TeCoA-2 & 0.037 $\pm$ 0.076 & 0.106 $\pm$ 0.127 & 0.161 $\pm$ 0.145 & 0.245 $\pm$ 0.160 \\
     & TeCoA-4 & 0.036 $\pm$ 0.077 & 0.102 $\pm$ 0.131 & 0.156 $\pm$ 0.148 & 0.239 $\pm$ 0.161 \\
     & FARE-2 & 0.077 $\pm$ 0.119 & 0.162 $\pm$ 0.152 & 0.218 $\pm$ 0.152 & 0.302 $\pm$ 0.152 \\
     & FARE-4 & 0.054 $\pm$ 0.097 & 0.130 $\pm$ 0.143 & 0.184 $\pm$ 0.146 & 0.265 $\pm$ 0.150 \\
    \bottomrule
    \end{tabular}
\end{table}

\subsection{Further Visualizations}
\label{sec:at_att_maps}

\subsubsection{Contrast of Attribution Maps}

In this section, we provide additional qualitative examples of attribution maps. While the examples of the Figure \ref{fig:att_map} were randomly sampled from the FARE-4 and TeCoA-4 decoders, we further explore qualitative differences using an alternative sampling strategy.

\begin{figure}
  \centering
  \begin{subfigure}{0.49\linewidth}
    \centering
    \includegraphics[height=0.25\textheight,width=\linewidth]{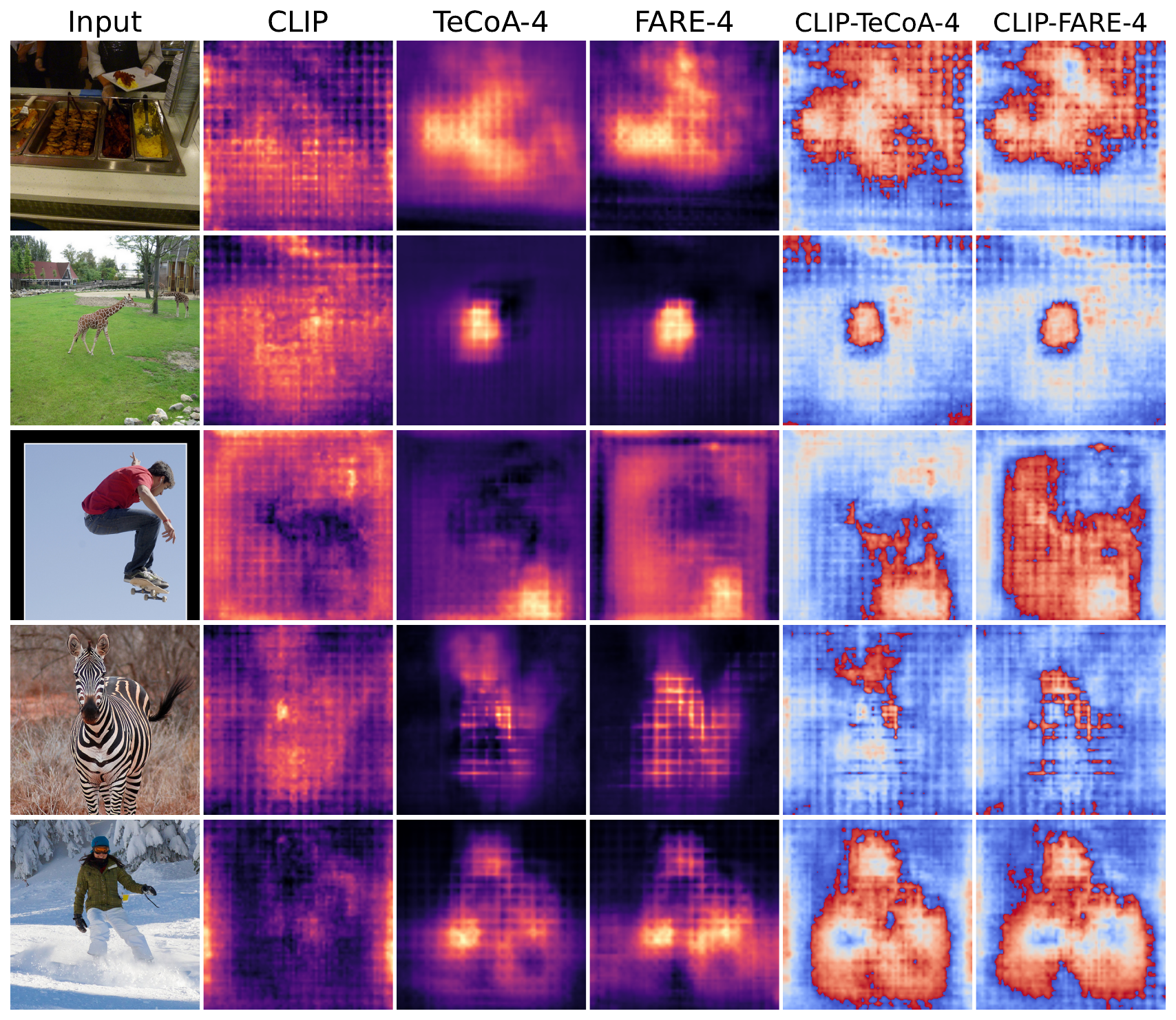}
  \end{subfigure}
  \hspace{0.0001\linewidth}
  \begin{subfigure}{0.49\linewidth}
  \centering
    \includegraphics[height=0.25\textheight,width=\linewidth]{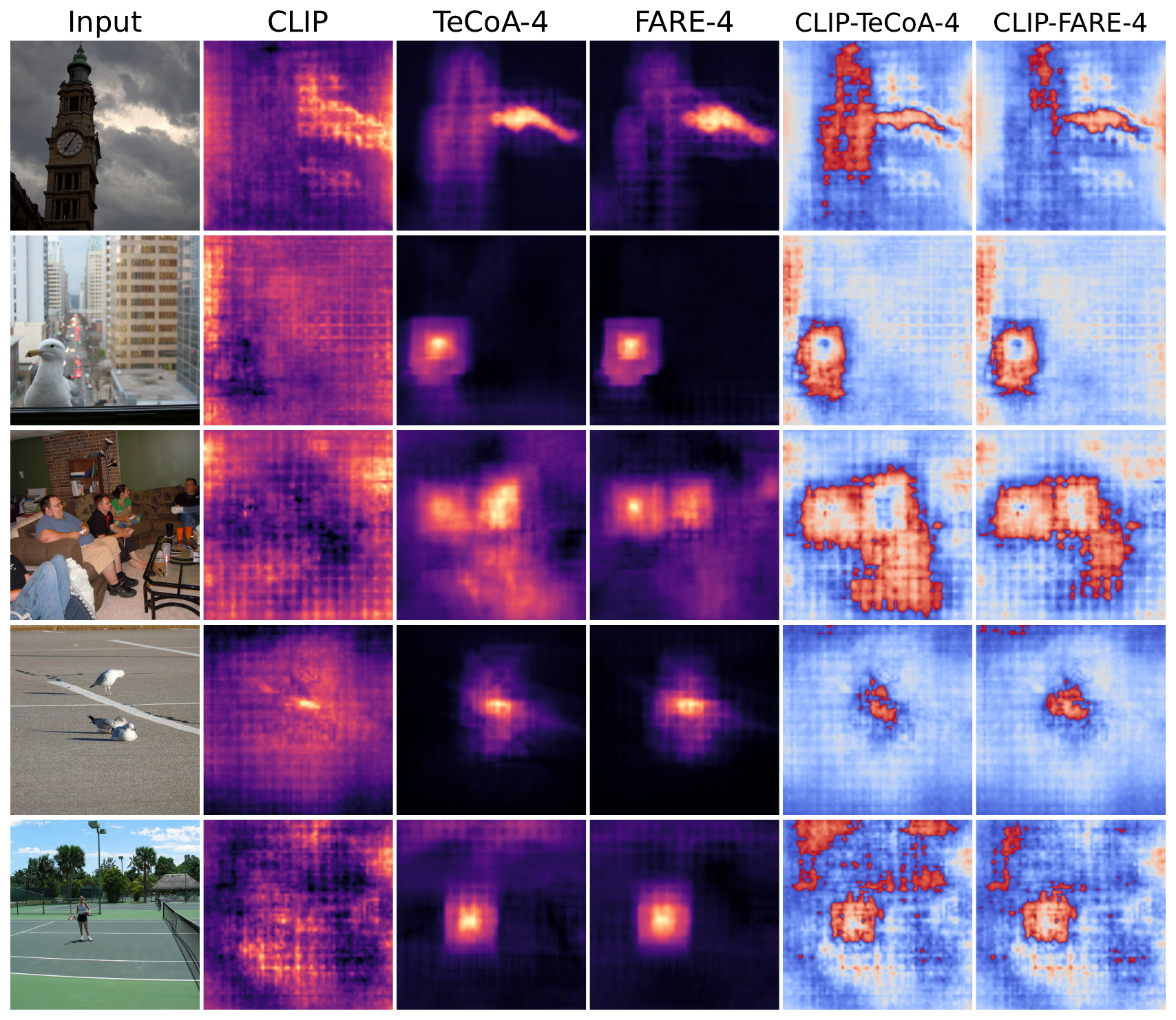}
  \end{subfigure}
  \caption{Attribution maps of 10 selected images from the test dataset by high distributional differences measured by JSD (Jensen-Shannon-Divergence). From left to right: input image, CLIP, TeCoA-4, FARE-4, difference between CLIP and TeCoA-4, and difference between CLIP and FARE-4. All the attribution maps are generated from Subject 01.}
    \label{fig:at_att_map_jsd}
\end{figure}

Specifically, Figure~\ref{fig:at_att_map_jsd} presents a subset of test images selected based on the Jensen--Shannon divergence (JSD), which we use to identify samples exhibiting highly divergent attribution distributions. Let $(\mathcal{X}, \mathcal{F})$ be a measurable space, and let $P$ and $Q$ be two probability distributions on $\mathcal{X}$ such that $P \ll M$ and $Q \ll M$, where $M = \frac{1}{2}(P + Q)$ denotes their mixture distribution. The \emph{Jensen--Shannon divergence} between $P$ and $Q$ is defined as
\begin{equation}
\mathrm{JSD}(P \,\|\, Q)
\;=\;
\frac{1}{2} D_{\mathrm{KL}}(P \,\|\, M)
\;+\;
\frac{1}{2} D_{\mathrm{KL}}(Q \,\|\, M),
\end{equation}
where $D_{\mathrm{KL}}$ denotes the Kullback--Leibler divergence:
\begin{equation}
D_{\mathrm{KL}}(P \,\|\, Q)
\;=\;
\int_{\mathcal{X}} \log\!\left(\frac{dP}{dQ}\right) dP.
\end{equation}

In our setting, $P$ and $Q$ correspond to normalized attribution maps produced by different models for the same input. 
By selecting samples with high JSD, we emphasize cases where the spatial distribution of attribution differs most substantially across models.

Qualitatively, these samples reveal a consistent pattern. The standard CLIP decoder tends to produce spatially diffuse attribution maps, whereas TeCoA and FARE decoders exhibit more spatially concentrated responses aligned with objects in the scene. Importantly, although no explicit constraint was imposed to encourage either spatial concentration or diffusion, a consistent qualitative pattern nevertheless emerges across models.

Furthermore, consistent with the observations discussed in the main text, the standard CLIP decoder often assigns non-negligible attribution mass to background regions, while robust CLIP variants yield more localized activation patterns. These differences become particularly pronounced in the high-divergence samples selected by JSD.

\subsubsection{Comparison between Adversarial Strengths}
\begin{figure}
  \centering
  \begin{subfigure}{0.49\linewidth}
    \centering
    \includegraphics[height=0.25\textheight,width=\linewidth]{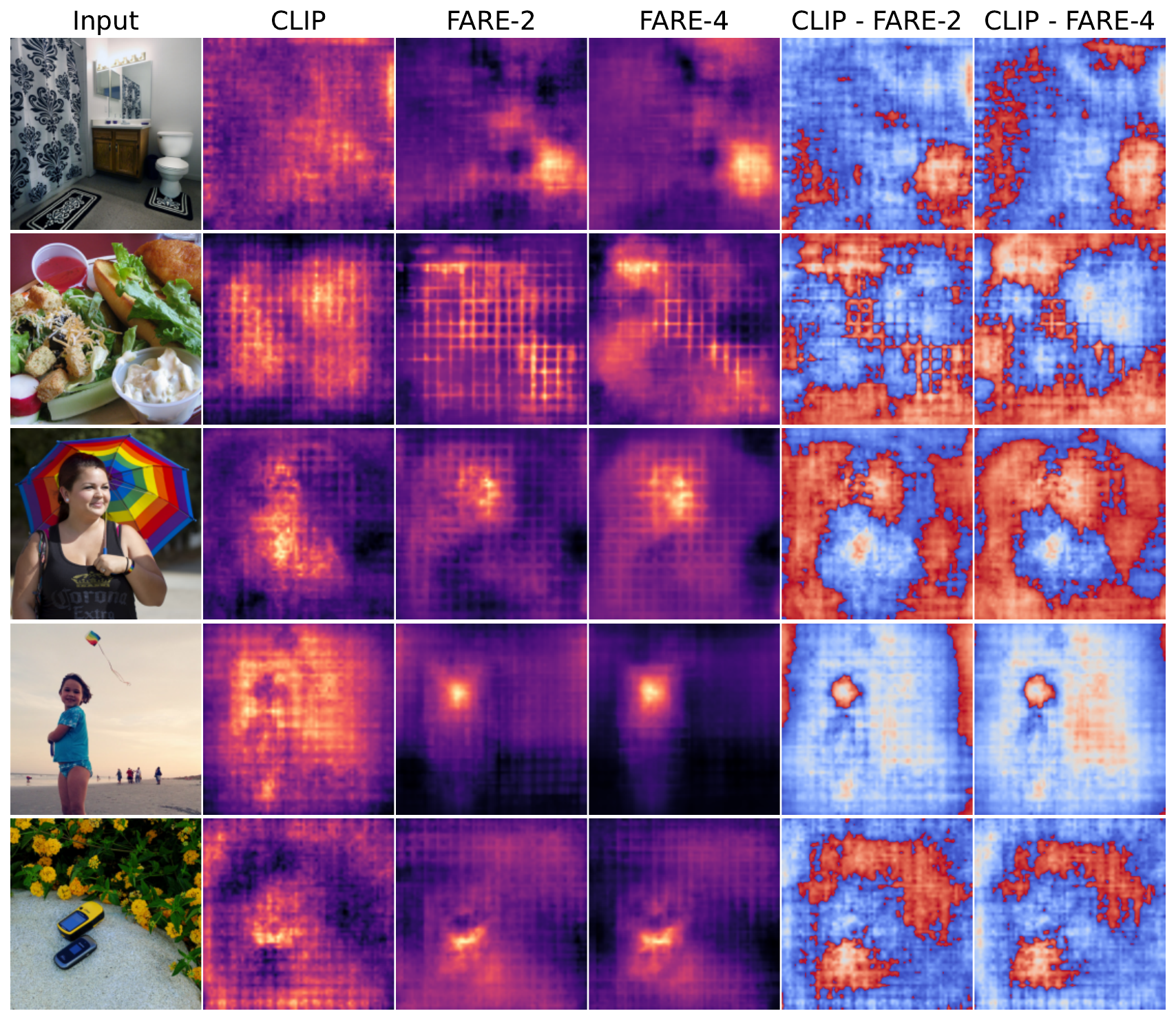}
  \end{subfigure}
  \hspace{0.0001\linewidth}
  \begin{subfigure}{0.49\linewidth}
  \centering
    \includegraphics[height=0.25\textheight,width=\linewidth]{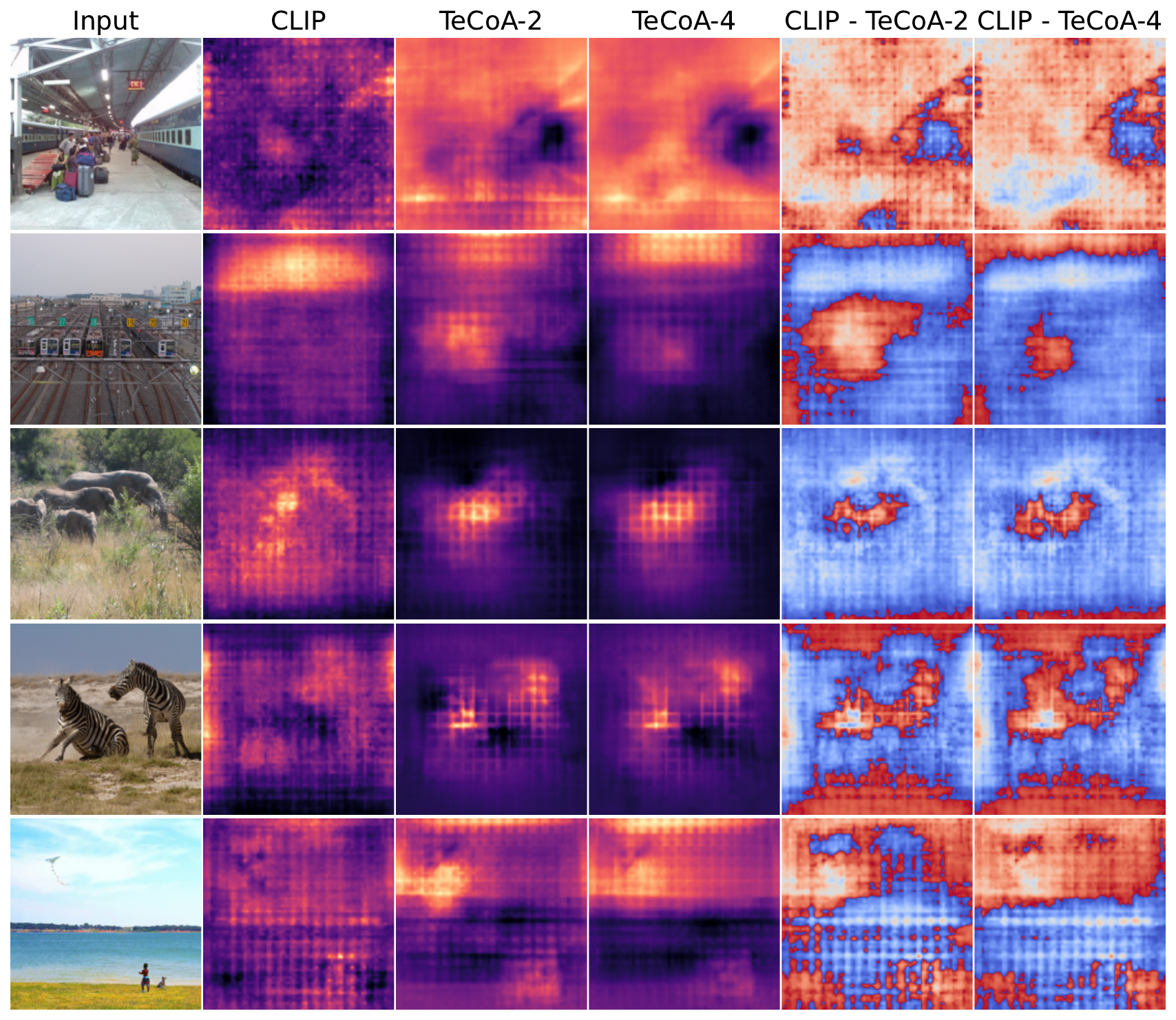}
  \end{subfigure}
  \caption{ (Left) Attribution maps of 5 randomly selected images from the test dataset. FARE-2 and FARE-4 were selected for the comparison between different adversarial training strengths. From left to right: input image, CLIP, FARE-2, FARE-4, difference between CLIP and FARE-2, and difference between CLIP and FARE-4. (Right) Attribution maps of 5 randomly selected images from the test dataset. TeCoA-2 and TeCoA-4 were selected for the comparison between different adversarial training strengths. From left to right: input image, CLIP, TeCoA-2, TeCoA-4, difference between CLIP and TeCoA-2, and difference between CLIP and TeCoA-4. All the attribution maps are generated from Subject 01.}
  \label{fig:at_att_map_eps2}
\end{figure}

Meanwhile, another interesting question is about the effect of adversarial training strengths (eps) used for adversarial training of the target representations. Therefore, we present the attribution map comparison among CLIP, moderately robust models ($\epsilon=2$), and strongly robust models ($\epsilon=4$) in the Figure~\ref{fig:at_att_map_eps2}. While the contrast between CLIP and moderately robust models is more noticeable, we observe the salient activation regions between moderately robust models and strongly robust models remain qualitatively similar. The examples are selected randomly from test dataset.

\section{Broader Impact}

This research aims to contribute to the intersection of neuroscience and AI security, particularly in brain decoding and the adversarial robustness of deep learning models. While brain decoding has potential applications in brain--computer interfaces (BCIs), our work focuses on improving the reliability and performance of brain decoding pipelines. At the same time, brain decoding technologies raise important ethical considerations. Neural data may contain sensitive and personal information, introducing risks related to privacy, consent, and potential misuse. It is therefore critical that research in this area is conducted with appropriate safeguards. In this work, we use only publicly available, de-identified datasets and do not involve any new data collection from human participants. Although our study focuses on methodological improvements rather than direct real-world deployment, we emphasize that future progress in this field should be accompanied by careful consideration of ethical, legal, and societal implications, including strong standards for data protection and responsible use.


\end{document}